\documentclass[10pt]{article} % For LaTeX2e
\usepackage[accepted]{tmlr}
\usepackage{amsmath,amsfonts,bm}

\def\eqref#1{equation~\ref{#1}}
\def\1{\bm{1}}

\DeclareMathAlphabet{\mathsfit}{\encodingdefault}{\sfdefault}{m}{sl}
\SetMathAlphabet{\mathsfit}{bold}{\encodingdefault}{\sfdefault}{bx}{n}

\usepackage{hyperref}
\usepackage{url}

\author{\name Leona Hennig \email l.hennig@ai.uni-hannover.de \\
      \addr Institute of Artificial Intelligence\\
      Leibniz University Hanover
      \AND
      \name Marius Lindauer \email m.lindauer@ai.uni.hannover.de \\
      \addr Institute of Artificial Intelligence \\
      \addr Leibniz University Hanover \\
      \addr L3S Research Center}

\usepackage{hyperref}
\usepackage{url}
\usepackage{latexsym}
\usepackage{amssymb}
\usepackage{amsmath}
\usepackage{float}
\usepackage{amsthm}
\usepackage{booktabs}
\usepackage{enumitem}
\usepackage{graphicx}
\usepackage{color}
\usepackage{float}
\usepackage{wrapfig}
\usepackage{algorithm}
\usepackage{algpseudocode}

\usepackage{amsmath}
\usepackage{amssymb}
\usepackage{amsfonts}
\usepackage{xcolor}
\usepackage{subcaption}
\usepackage{multirow}

\usepackage{booktabs}

\usepackage{microtype}
\usepackage{mathtools}

\usepackage{caption}
\usepackage{hyperref}
\usepackage{url}
\usepackage{microtype}
\usepackage{graphicx}
\usepackage{booktabs}
\usepackage{array}
\usepackage{xcolor}
\usepackage{marginnote}
\usepackage{amsmath}
\usepackage{mathrsfs}
\usepackage{amssymb}
\usepackage{graphicx} 
\usepackage{caption}
\usepackage{tabularx}
\usepackage{enumitem}
\numberwithin{equation}{section}

\title{Leveraging AutoML for Sustainable Deep Learning: A Multi-Objective HPO Approach on Deep Shift Neural Networks}

\def\month{07}  % Insert correct month for camera-ready version
\def\year{2025} % Insert correct year for camera-ready version
\def\openreview{\url{https://openreview.net/forum?id=vk7b11DHcW}} % Insert correct link to OpenReview for camera-ready version

\begin{document}

\maketitle

\begin{abstract}
Deep Learning (DL) has advanced various fields by extracting complex patterns from large datasets.
However, the computational demands of DL models pose environmental and resource challenges. Deep Shift Neural Networks (DSNNs) present a solution by leveraging shift operations to reduce computational complexity at inference.
Compared to common DNNs, DSNNs are still less well understood and less well optimized. 
By leveraging AutoML techniques, we provide valuable insights into the potential of DSNNs and how to design them in a better way.
\textcolor{black}{
We focus on image classification, a core task in computer vision, especially in low-resource environments.}
Since we consider complementary objectives such as accuracy and energy consumption, we combine state-of-the-art multi-fidelity (MF) hyperparameter optimization (HPO) with multi-objective optimization to find a set of Pareto optimal trade-offs on how to design DSNNs.
Our approach led to significantly better configurations of DSNNs regarding loss and emissions compared to default DSNNs. This includes simultaneously increasing performance by about 20\% and reducing emissions, in some cases by more than 60\%.
Investigating the behavior of quantized networks in terms of both emissions and accuracy, our experiments reveal surprising model-specific trade-offs, yielding the greatest energy savings.
For example, in contrast to common expectations, quantizing smaller portions of the network with low precision can be optimal with respect to energy consumption while retaining or improving performance.
We corroborated these findings across multiple backbone architectures, highlighting important nuances in quantization strategies and offering an automated approach to balancing energy efficiency and model performance.
\end{abstract}

\section{Introduction}
\label{Introduction}

Deep Learning (DL) is a promising approach to extracting information from large datasets with complex structures. This includes performing computations in IoT environments and on edge devices~\citep{DBLP:journals/network/LiOD18,DBLP:journals/pieee/ZhouCLZLZ19}, which can come with strict limitations on energy consumption.
This is especially true for the ever-increasing size and performance of such models due to the progress in science and industry. Running these models is not free of computational costs~\citep{DBLP:journals/pieee/SzeCYE17}, and minimizing this cost directly affects the environmental impact of a model~\citep{DBLP:journals/cacm/SchwartzDSE20}. 
Even if resource consumption should not be considered crucial in view of environmental impact, efficiently designed neural networks free up resources that can be used for other tasks, e.g., edge computing or computations for advanced driver assistance systems~\citep{DBLP:journals/corr/HowardZCKWWAA17}. 
With our approach, we contribute to DL by minimizing its environmental footprint and allowing applications in low-resource settings.

In this context, we focus on image classification, one of the core tasks in computer vision and particularly relevant for low-power applications. Image classification is widely common in scenarios where resources are limited, such as edge computing, automated driving, and industrial production environments. These applications require fast, accurate, and energy-efficient models that can process visual information in real time. Because it is well understood and broadly benchmarked, image classification enables reproducible evaluation of architectural and optimization strategies with minimal confounding of task-specific variables.

Of particular interest to us are Deep Shift Neural Networks (DSNNs), which offer great potential to reduce power consumption compared to traditional DL models, e.g., by reducing the inference time by a factor of 4~\citep{DBLP:conf/cvpr/ElhoushiCSTL21}. 
Instead of expensive floating-point arithmetic, they leverage cheap shift operations - specifically, bit shifting - as the computational unit, which boosts efficiency by replacing costly multiplication operations in convolutional networks.
Although DSNNs offer great promise, so far, different design decisions, including training hyperparameters and shift architecture, have not been well studied, and there is little knowledge about their full potential.
We suspect that the configuration of DSNNs has a huge impact on both performance and computational efficiency.

One of the key challenges with DSNNs is determining the appropriate level of precision for shift operations to minimize quantization errors without excessively increasing the computational load. 
To address this challenge, we propose to apply automated machine learning (AutoML) to DSNNs to find their optimal configuration. 
This is achieved by hyperparameter optimization (HPO)~\citep{bischl-dmkd23a} and a neural architecture search on a macro-level~\citep{elsken-jmlr19a}. 
Integrating multi-fidelity (MF) and multi-objective optimization (MO) techniques~\citep{Belakaria2020-re} facilitates an optimal exploration of the configuration space that trades off predictive performance and energy consumption~\citep{Deb2014}. 
To this end, we extended the SMAC3 approach~\citep{lindauer-jmlr22a}, a state-of-the-art HPO package~\citep{eggensperger-neuripsdbt21a}, so that its MO implementation effectively balances the trade-off between achieving high predictive accuracy and minimizing energy consumption. Using tools such as CodeCarbon~\citep{DBLP:journals/corr/abs-1910-09700,DBLP:journals/corr/abs-1911-08354} during the training and evaluation phases provides insight into the energy consumption and carbon emissions associated with each model configuration.
The MF aspect allows for the efficient use of computational resources by evaluating configurations at varying levels of approximation. Our work is in the spirit of Green AutoML \citep{tornede-jair23a}, considering efficient AutoML to gain insight into the design of efficient DSNNs.

\paragraph{Contributions}
Overall, we contribute to Green AutoML w.r.t. DSNNs by:
\begin{enumerate}
    \item Specifying the first configuration space tailored to DSNNs, which is efficiently optimized by a combination of multi-objective and multi-fidelity AutoML approaches;
    \item Empirically exploring how specific design choices in DSNNs lead to different trade-offs between accuracy and energy efficiency, enabling stakeholders and researchers to leverage these findings to develop energy-efficient applications that maintain high computational accuracy; and
    \item Identifying specific configurations of DSNNs that surpass the baseline results in both dimensions of the performance-efficiency optimization problem.
\end{enumerate}

The corresponding repository is available at \url{https://github.com/automl/Auto-DSNN}.

\section{Related Work}
\label{sec:related_work}
Both multi-fidelity optimization  (MF)~\citep{bischl-dmkd23a} and multi-objective optimization (MO) ~\citep{moraleshernandez-air22a} for AutoML have gotten a lot of traction in recent years.
The combined integration of multi-fidelity multi-objective optimization (MFMO) has seen some advancements to enhance the efficiency of model training while minimizing environmental impact. \cite{Belakaria2020-re} proposed an acquisition function based on output space entropy search for multi-fidelity multi-objective Bayesian optimization (MFMO-BO-OSES). Their method addresses the exploration-exploitation dilemma by prioritizing the acquisition of data points that significantly reduce the entropy of the Pareto front. This approach enables more strategic sampling decisions and leverages lower-fidelity evaluations to approximate the Pareto front effectively, aligning with the sustainability goals of Green AutoML. Similarly, \cite{schmucker-metalearn20a} considered a combination of multi-objective and multi-fidelity optimization but focused on fairness as the second objective. With our MFMO approach, we contribute an algorithm tailored specifically for performing efficient HPO tasks using BO, directly minimizing emissions in the process.

A well-explored technique for reducing the computational complexity of neural networks themselves is network quantization. It involves lowering the precision of weights and activations, which decreases the model's memory footprint and accelerates inference. Works by  \cite{courbariaux-nips15a} and  \cite{DBLP:journals/corr/RastegariORF16}, for example, have demonstrated that techniques such as BinaryConnect and XNOR-Net not only reduce computational requirements but also maintain near-state-of-the-art performance, underscoring the potential of quantization to balance performance with computational efficiency. This has also been transferred into the field of LLMs, where 1-Bit transformer architectures are used to address the challenges of increasing model size~\citep{DBLP:journals/corr/abs-2310-11453}.
Strongly related to network quantization, Deep Shift Neural Networks (DSNNs), proposed by \cite{DBLP:conf/cvpr/ElhoushiCSTL21}, represent an advancement towards quantization combined with efficient operators. DSNNs employ bitwise shift operations instead of traditional multiplications, thus further reducing the computational overhead and power consumption. This innovation is particularly crucial for deploying Deep Learning models in power-sensitive or resource-constrained environments, further contributing to the environmental sustainability of AI technologies.

DSNNs offer a more balanced trade-off between computational efficiency and accuracy compared to binary methods like BinaryConnect and XNOR-Net by replacing multiplications with bitwise shifts, which are more power-efficient yet maintain greater precision, reducing the accuracy loss typically associated with binary quantization. This allows DSNNs to achieve better performance in resource-constrained environments while still minimizing computational overhead~\citep{DBLP:conf/cvpr/ElhoushiCSTL21}.

AdderNet~\citep{DBLP:conf/cvpr/ChenWXSX0X20} is another approach for reducing the amount of computationally expensive multiplications during network training and inference by using the $\ell_1$-norm distance between input and filter vectors to compute activations. There are efforts in mixed-precision quantization, where different bit-widths are assigned to different layers or channels of the network. It allows for higher precision where necessary and lower precision where possible, optimizing the trade-off between accuracy and efficiency~\citep{DBLP:journals/corr/abs-2407-01054}. Similarly, ternary quantization (TTQ) constrains the weights to three discrete values: $\{-1, 0, +1\}$. To mitigate the accuracy loss from reduced precision, TTQ introduces learned scaling factors, allowing networks to maintain performance comparable to their full-precision counterparts while significantly reducing memory and power usage~\citep{DBLP:journals/tist/RokhAK23}.
HPO combined with DSNNs further optimizes both performance and resource efficiency by tailoring shift depths and quantization strategies, allowing for fine-tuned control over energy consumption and accuracy in constrained environments, making them ideal for mixed-precision and quantization-aware neural network applications.

\section{Background}
\label{background}
The following chapter introduces foundational concepts for our approach.
\subsection{Deep Shift Neural Networks}
A Deep Shift Neural Network (DSNN) is an approach to reduce the computational and energy demands of Deep Learning~\citep{DBLP:conf/cvpr/ElhoushiCSTL21}. DSNNs achieve a considerable reduction in latency time by simplifying the network architecture such that they replace the traditional multiplication operations in neural networks with bit-wise shift operations and sign flipping, making DSNNs suitable for computing devices with limited resources.
There are two methods for training DSNNs~\citep{DBLP:conf/cvpr/ElhoushiCSTL21}: DeepShift-Q (Quantization) and DeepShift-PS (Powers of two and Sign). 
DeepShift-Q involves training regular weights constrained to powers of 2 by quantizing weights to their nearest power of two during both forward and backward passes. In DeepShift-Q, the weights are quantized to powers of two by rounding the logarithm of the absolute weights to the nearest integer. This process simplifies the weight representation and ensures compatibility with bitwise shift operations. The sign is then applied to preserve the original weight polarity.
DeepShift-PS directly includes the values of the shifts and sign flips as trainable hyperparameters, offering finer control over weight adaptation. This approach removes the need for explicit rounding during training, potentially leading to improved precision at the cost of additional parameter updates.

The DeepShift-Q approach obtains the sign matrix \(S\) from the trained weight matrix \(W\) as \mbox{\( S = \mathit{sign}(W) \)}. 
The power matrix \(P\) is the base-2 logarithm of \(W\)'s absolute values, i.e., \mbox{\( P = \log_{2}(|W|) \)}. 
After rounding \(P\) to the nearest power of two, \(  \tilde{P}  = \mathit{round}(P) \), the quantized weights \(\tilde{W}_q\) are calculated by applying the sign \(S\):
\begin{equation}
    \tilde{W}_q = \mathit{flip}(2^{ \tilde{P} }, S)\:.
\end{equation}
The DeepShift-PS approach optimizes neural network weights by directly adapting the shift (\( \tilde{P} \)) and sign (\( \tilde{S} \)) values. 
The shift matrix \( \tilde{P} \) is obtained by rounding the base-2 logarithm of the weight values, \mbox{\( \tilde{P} = \mathit{round}(P) \)}, and the sign flip \( \tilde{S} \) is computed as \mbox{\( \tilde{S} = \mathit{sign}(\mathit{round}(S)) \).} 
Weights are calculated as
\begin{equation}
    \tilde{W}_{\mathit{ps}} = \mathit{flip}(2^{\tilde{P}}, \tilde{S})\:,
\end{equation} where the sign flip operation \( \tilde{S} \) assigns values of \(-1\), \(0\), or \(+1\) based on \( S \). 
Directly training shift and sign values could allow for more precise control in optimizing the network's computational efficiency by reducing mathematical imprecision.
However, training the floating point weights and only rounding them during the forward and backward passes might increase the precision and reduce the error in training the weights.

\subsection{Hyperparameter Optimization}
\label{hpo}
The increasing complexity of Deep Learning algorithms enhances the need for automated hyperparameter optimization (HPO) to increase model performance~\citep{bischl-dmkd23a}. 
Consider a dataset \( \mathcal{D} = \{(x_i,y_i)\}_{i=1}^{N} \in \mathbb{D} \subset \mathcal{X} \times \mathcal{Y} \), where \( \mathcal{X} \) is the instance space and \( \mathcal{Y} \) is the target space, and a hyperparameter configuration space \( \Lambda = \{\lambda_1, \ldots, \lambda_L\} \), \( L \in \mathbb{N} \). 
In our work, 
\( \mathcal{M} \) denotes the space of possible DSNN models. The dataset \( \mathcal{D} \) is split into disjoint training, validation, and test sets: \( \mathcal{D}_{\textit{train}}, \mathcal{D}_{\textit{val}}, \) and \( \mathcal{D}_{\textit{test}} \) respectively. An algorithm \( \mathcal{A} : \mathbb{D} \times \Lambda \rightarrow \mathcal{M} \) trains a model \( M \in \mathcal{M} \), instantiated with a configuration of \( L \) hyperparameters sampled from \( \Lambda \), on the training data \( \mathcal{D}_{\textit{train}} \).  The performance of the algorithm is assessed via an expensive-to-evaluate loss function \( \mathcal{L} : \mathcal{M}_\lambda \times \mathbb{D} \rightarrow \mathbb{R} \), which involves both the training on \( \mathcal{D}_{\textit{train}} \) and the evaluation of the model on \( \mathcal{D}_{\textit{val}} \). The optimization objective of HPO is to find configurations \( \lambda^* \in \Lambda \) with minimal validation loss \( \mathcal{L} \):
\begin{equation}
\lambda^* \in \arg\min_{\lambda \in \Lambda} \mathcal{L}\big(\mathcal{A}(\mathcal{D}_{\textit{train}},\lambda), \mathcal{D}_{\textit{val}}\big).
\end{equation}
Finally, the model's final performance is assessed on \(\mathcal{D}_{\textit{test}}\).

\subsection{Bayesian Optimization}
\label{bo}

For a given dataset, Bayesian Optimization (BO) for HPO is a strategy for global optimization of black-box loss functions \mbox{$\mathcal{L}:\mathcal{M}_\lambda\times\mathbb{D}\xrightarrow{}\mathbb{R}$} that are expensive to evaluate~\citep{DBLP:journals/jgo/JonesSW98}.

BO uses a probabilistic surrogate model $\mathcal{S}$ to approximate the loss function, commonly given by a Gaussian Process or a Random Forest~\citep{DBLP:books/lib/RasmussenW06,hutter-lion11a,shahriari-ieee16a}. 
An acquisition function $\alpha:\Lambda\xrightarrow{}\mathbb{R}$ guides the search for the next optimal evaluation points by balancing the exploration-exploitation trade-off, based on the set of previously evaluated configurations $\{(\lambda_1,\mathcal{L}(M_{\lambda_1},\cdot)),...,(\lambda_{m-1},\mathcal{L}(M_{\lambda_{m-1}},\cdot))\}$ at time $m$.
Common choices for acquisition functions include expected improvement (EI)~\citep{DBLP:journals/jgo/JonesSW98} since it calculates the expected improvement in the objective function value and guides the search towards regions where improvements are most likely.

Entropy-based methods like Entropy Search (ES)~\citep{hennig-jmlr12a} and Predictive Entropy Search (PES) \citep{hernandez-nips14a} aim to reduce the entropy of the posterior distribution of the maximizer, focusing on information-rich regions.
The Knowledge Gradient (KG)~\citep{frazier-jc09a} offers a strategy for maximizing the expected improvement of the objective, considering all potential outcomes, valuable in scenarios with noisy measurements. 
BO is particularly well-suited for hyperparameter optimization in Deep Learning, where evaluating the performance of a model configuration can be computationally expensive because of the training of each configuration. 
BO is sample-efficient in evaluating $\mathcal{L}$ on only a few configurations. 

\subsection{Multi-Fidelity Optimization}
\label{multi-fidelity}

To reduce the cost of fully training multiple DSNN configurations, we employ a multi-fidelity (MF) approach~\citep{DBLP:journals/jmlr/LiJDRT17}, which is a common strategy in AutoML to navigate the trade-off between performance and approximation error~\citep{DBLP:books/sp/HKV2019}. 
MF approaches train cheap-to-evaluate proxies of black-box functions following different heuristics, e.g., allocating a small number of epochs to many configurations in the beginning and training the best-performing ones on an increasing number of epochs. 
Formally, we define a space of fidelities $\mathcal{F}$ and aim to minimize a function $F\in\mathcal{F}$ \citep{DBLP:journals/jair/KandasamyDOSP19}:
\begin{equation}
    \min_{\lambda\in\Lambda} F(\lambda)\, .
\end{equation}
We approximate $F\in\mathcal{F}$, using a series of lower-fidelity, i.e., less expensive approximations 
$\{f(\lambda)_1,\ldots,f(\lambda)_j=F(\lambda)\}$, where $j$ denotes the total number of fidelity levels.
The target function $F\in\mathcal{F}$ corresponds to the loss function $\mathcal{L}$ of HPO and BO.
The allocated resources for evaluating a model's performance at various fidelities are referred to as a budget, e.g., training a DNN for only $n\in\mathbb{N}$ epochs instead of until convergence. 
MF typically assumes that the highest fidelities approximate the black-box function best. The longer a model is trained, the more accurate its approximation of an underlying function gets.

\subsection{Multi-Objective Optimization}
Multi-objective optimization (MO) addresses problems involving multiple, often competing, objectives. This approach is used in scenarios where trade-offs between two or more conflicting objectives must be navigated, such as in the context of DSNNs, enhancing accuracy alongside reducing energy consumption. MO aims to identify Pareto optimal solutions~\citep{Deb2014}. New points are added based on the current observation dataset \( \mathcal{D}_\mathit{obs} = \{(\lambda_1,\mathcal{L}(M_{\lambda_1},\cdot), \ldots, (\lambda_n,\mathcal{L}(M_{\lambda_m},\cdot)\} \) at time $m+1$. These points augment the surface formed by the non-dominated solution set \( D^\star_n \), which satisfies the condition for $d$ objective variables and a loss function $\mathcal{L} = (\mathcal{L}_1,\ldots,\mathcal{L}_d)$, where $\mathcal{L}_k$ corresponds to the loss regarding objective $k$~\citep{DBLP:journals/corr/abs-1905-02370}: 
\begin{align}
\forall \lambda, (\lambda,\mathcal{L}(\lambda)) \in \mathcal{D}^\star_n \subset \mathcal{D}_n,\: (\lambda',\mathcal{L}(\lambda')) \in \mathcal{D}_n \exists k \in \{1,\ldots,d\}:\: \mathcal{L}_k(\lambda) \leq \mathcal{L}_k(\lambda').
\end{align}
W.l.o.g., we assume the minimization of all objectives.
The observation dataset $\mathcal{D}_\mathit{obs}$ is iteratively updated to search for solutions that approximate the Pareto front.

\section{Approach}
\label{approach}
Our goal is to provide deeper insights into the structure of Deep Shift Neural Networks (DSNNs). We optimize these networks with respect to both performance and efficiency, highlighting how their hyperparameters influence the optimization process and resulting trade-offs.

\subsection{Configuration Space Exploration}
\label{subsec:configuration_space}

The foundation of our approach lies in defining and exploring a robust configuration space tailored specifically to DSNNs. This space includes a range of hyperparameters that influence the network's performance and energy efficiency. Key hyperparameters under consideration include:

\begin{enumerate}
    \item Shift Depth: determines the number of network layers converted to employ shift operations, replacing conventional floating point operations and thereby reducing computational overhead.
    \item Shift Type: selects the method of shift operation, either quantization (DeepShift-Q) or direct training of shifts (DeepShift-PS), impacting the network's training dynamics and inference efficiency.
    \item Bit Precision for Weights and Activations: influences the network's accuracy and the granularity of its computations, affecting both performance and power consumption.
    \item Rounding Type: affects how weight adjustments are handled during training, with options for deterministic or stochastic rounding, each offering trade-offs in terms of stability and performance.
\end{enumerate}
Table \ref{table:model_config} details the configuration space for a ResNet20 model adapted for DSNNs, outlining the range and default values of each hyperparameter considered in our study.

% An illustration of how the investigated hyperparameters change the weight calculation in the network can be found in Figure \ref{fig:shifts1}.

% \begin{figure*}[htbp]
%   \centering
%   \includegraphics[width=0.9\textwidth, trim={0cm 0cm 0cm 5cm}, clip]{ecai-template/Unbenannte Präsentation (6).pdf}
%   \caption{Computational-wise comparison of different architectures specifying the computation of the weights. The use of shift type DeepShift-Q or DeepShift-PS determines whether the weights are calculated the standard way and rounded only during the forward and backward pass, with precision controlled by the number of weight bits, or whether the bit-shift and sign-flip operations are treated as directly trainable parameters. While the shift depth determines the number of shift layers and the number of activation bits determines the precision of the activation representations, the shift type and number of weight bits also significantly influence the accuracy of the training.}
%   \label{fig:shifts1}
% \end{figure*}

\subsection{Multi-Fidelity Multi-Objective Optimization Framework}
\label{subsec:mfmo_framework}

To computationally enhance DSNNs via AutoML, we employ multi-fidelity optimization (MF), see Section~\ref{background}. A well-known MF algorithm is successive halving~\citep{jamieson-aistats16a}, where $n_c$ configurations are trained on an initial small budget $b_I$. 
It addresses the trade-off between $b_I$ and $n_c$, or between approximation error and exploration inherent in successive halving, using the HyperBand algorithm for MF.
HyperBand~\citep{DBLP:journals/jmlr/LiJDRT17} runs successive halving in multiple brackets, where each bracket provides a combination of $n_c$ and a fraction of the total budget per configuration so that they sum up to the total budget. 
We extend this to multi-fidelity multi-objective optimization (MFMO). We simultaneously address the accuracy of the model as well as its energy consumption using a two-dimensional objective function:
\begin{equation}
    \mathcal{L}_{\text{MO}}:\Lambda\xrightarrow{}\mathbb{R}^2,\quad \mathcal{L}_{\text{MO}}(\lambda) = \big(\mathcal{L}_{\text{loss}}(\lambda), \mathcal{L}_{\text{emission}}(\lambda)\big),
\end{equation} where, given a configuration $\lambda\in\Lambda$, $\mathcal{L}_{\text{loss}}(\lambda)$ aims to minimize the loss, enhancing the model's accuracy, and $\mathcal{L}_{\text{emission}}(\lambda)$ seeks to minimize the energy consumption during training and inference, promoting environmental sustainability.
We aim to solve the following optimization problem:
\begin{equation}
    \arg\min_{\lambda\in\Lambda} \mathcal{L}_{\text{MO}}(\lambda)\, .
\end{equation}

\begin{table}[ht]
\centering
\caption{Configuration search space of ResNet20. The first half includes commonly used training hyperparameters for deep learning, whereas the second half is specific to DSNNs.}
\label{table:model_config}
\setlength{\tabcolsep}{6pt} % Adjust column spacing as needed
\renewcommand{\arraystretch}{1.2} % Adjust row spacing
\begin{tabular}{lcc}
\toprule
\textbf{Hyperparameter} & \textbf{Search Space} & \textbf{Default} \\
\midrule
\multicolumn{3}{c}{Standard Training Hyperparameters} \\
\midrule
Batch Size & [32, 128] & 128 \\
Optimizer & \{SGD, Adam, Adagrad, Adadelta, RMSProp, RAdam, Ranger\} & SGD \\
Learning Rate & [0.001, 0.1] & 0.1 \\
Momentum & [0.0, 0.9] & 0.9 \\
Weight Decay & [1e-6, 1e-2] & 0.0001 \\
\midrule
\multicolumn{3}{c}{DSNN-Specific Hyperparameters} \\
\midrule
Weight Bits & [2, 8] & 5 \\
Activation Integer Bits & [2, 32] & 16 \\
Activation Fraction Bits & [2, 32] & 16 \\
Shift Depth & [0, 20] & 20 \\
Shift Type & \{Q, PS\} & PS \\
Rounding & \{deterministic, stochastic\} & deterministic \\
\bottomrule
\end{tabular}
\end{table}

Our approach leverages the MFMO framework to efficiently explore the configuration space using computationally inexpensive proxies for full training. This enables broader and more tractable exploration of the hyperparameter landscape within practical resource constraints.
The method is inherently architecture- and task-agnostic, requiring no modifications when applied to different models or datasets, aside from adjusting the DSNN’s quantization mechanism to align with the specific target architecture and task.

\subsection{Algorithmic Implementation of MFMO}
\label{subsec:algorithmic_implementation}

We use the ParEGO algorithm~\citep{DBLP:journals/tec/Knowles06} to compute Pareto optimal configurations. It transforms the multi-objective problem into a series of single-objective problems by introducing varying weights for the objectives in each iteration of HyperBand, thus optimizing a different scalarization per evaluation to approximate the Pareto front.
The resulting single-objective optimization function can then be evaluated in an MF setting. All configurations having survived a successive halving bracket are checked against the current Pareto front approximation and the Pareto set is updated if necessary.
The computational strategy initially involves computing a broad array of configurations and leveraging the successive halving method to efficiently narrow down the field to the most promising candidates.
We specifically target solutions that represent both extremes of the Pareto front, those that excel in one objective at the potential expense of the other, and configurations that provide a balanced compromise between the two objectives.
We included pseudocode of our algorithmic implementation in Algorithm \ref{algo:parego_dsnn} in the appendix.

% \subsection{Experimental Validation}
% \label{subsec:experimental_validation}

% In the next section, we systematically compare these optimized configurations against the default settings of the DSNNs. This comparison aims to quantify the improvements achieved through our MFMO approach, specifically in terms of accuracy and energy efficiency. By demonstrating substantial gains over the default configurations, we gain insights into what design choices impact those improvements and how we can make use of them for sustainable computations based on DSNNs.

\section{Experiments}
\label{experiments}

In the following section, we detail the setup and methodology used to evaluate our approach discussed in Section \ref{approach}, focusing on optimizing Deep Shift Neural Networks (DSNNs) through multi-fidelity, multi-objective optimization (MFMO), and extending the  DSNN objective function to multi-objective to compute a Pareto front for optimality regarding performance and efficiency. We discuss how our approach successfully navigates the model performance and environmental impact trade-offs. From this, we gain insights into DSNNs and how specific design choices might affect their performance. By identifying optimal configurations for both or either objectives, we draw conclusions about how the DSNN-specific hyperparameters in the network architecture interact with each other.

\subsection{Evaluation Setup}
\label{eval setup}
We train and evaluate our models on the CIFAR10 dataset~\citep{krizhevsky2009learning} and the Caltech101 dataset~\citep{DBLP:conf/cvpr/LiFP04}.
Our experiments are conducted on the NVIDIA A100, a widely used GPU, standard in research and industry for some time. Its computational capabilities and availability in many high-performance clusters make it a popular choice in the machine learning and deep learning communities. Using the A100, we ensure that our findings are broadly applicable and relevant to real-world scenarios, aligned with hardware commonly utilized for training and deploying advanced neural networks.

For hyperparameter optimization (HPO) with multi-fidelity optimization, we extend SMAC3~\citep{hutter-lion11a,lindauer-jmlr22a}, as well-known state-of-the-art HPO package~\citep{eggensperger-neuripsdbt21a}.
As a starting point, we chose the well-known ResNet20~\citep{DBLP:conf/cvpr/HeZRS16} architecture as used by \cite{DBLP:conf/cvpr/ElhoushiCSTL21}. Overall, this architecture is well understood and allows us to study DSNNs with few confounding factors. Additionally, we evaluate our approach using the well-known GoogLeNet~\citep{DBLP:conf/cvpr/SzegedyLJSRAEVR15} and MobileNetV2 architectures~\citep{sandler-cvpr18a}. We follow the model implementation of \cite{DBLP:conf/cvpr/ElhoushiCSTL21} to ensure comparability.
The configuration space is given in Table \ref{table:model_config}, for which we focus on the DSNN-specific hyperparameters (lower part of the table) and general training hyperparameters (upper part).
The fidelities are the number of epochs.

For multi-objective optimization, we aim to compute a Pareto front of optimal configurations for performance and energy consumption.
To incorporate the environmental impact into our HPO workflow, we use the CodeCarbon emissions tracker~\citep{DBLP:journals/corr/abs-1910-09700,DBLP:journals/corr/abs-1911-08354} to track carbon emissions from computational processes, in our case inference, by monitoring energy use and regional energy mix in kgCo2eq, grams of $\text{CO}_2$ equivalents. We use power consumption as a proxy for $\text{CO}_2$ emissions, acknowledging that external factors such as energy source and grid transmission losses are not directly controlled.
These emission values are incorporated into SMAC3 alongside DSNN's performance metric. The loss is $1-$ accuracy. 
%This integration helps balance DSNN performance with environmental sustainability, guiding the selection of efficient and eco-friendly hyperparameter configurations.

% The specific training budgets applied were 3, 7, 20, and 60 epochs (rounded up from the assigned budgets), respectively, allowing us to thoroughly explore the trade-offs between computational efficiency and model effectiveness across different stages of training.

\subsection{Results}

\subsubsection{Quantitative Results}

\begin{figure}
    \centering

    \begin{subfigure}[t]{0.47\textwidth}
        \includegraphics[width=\textwidth]{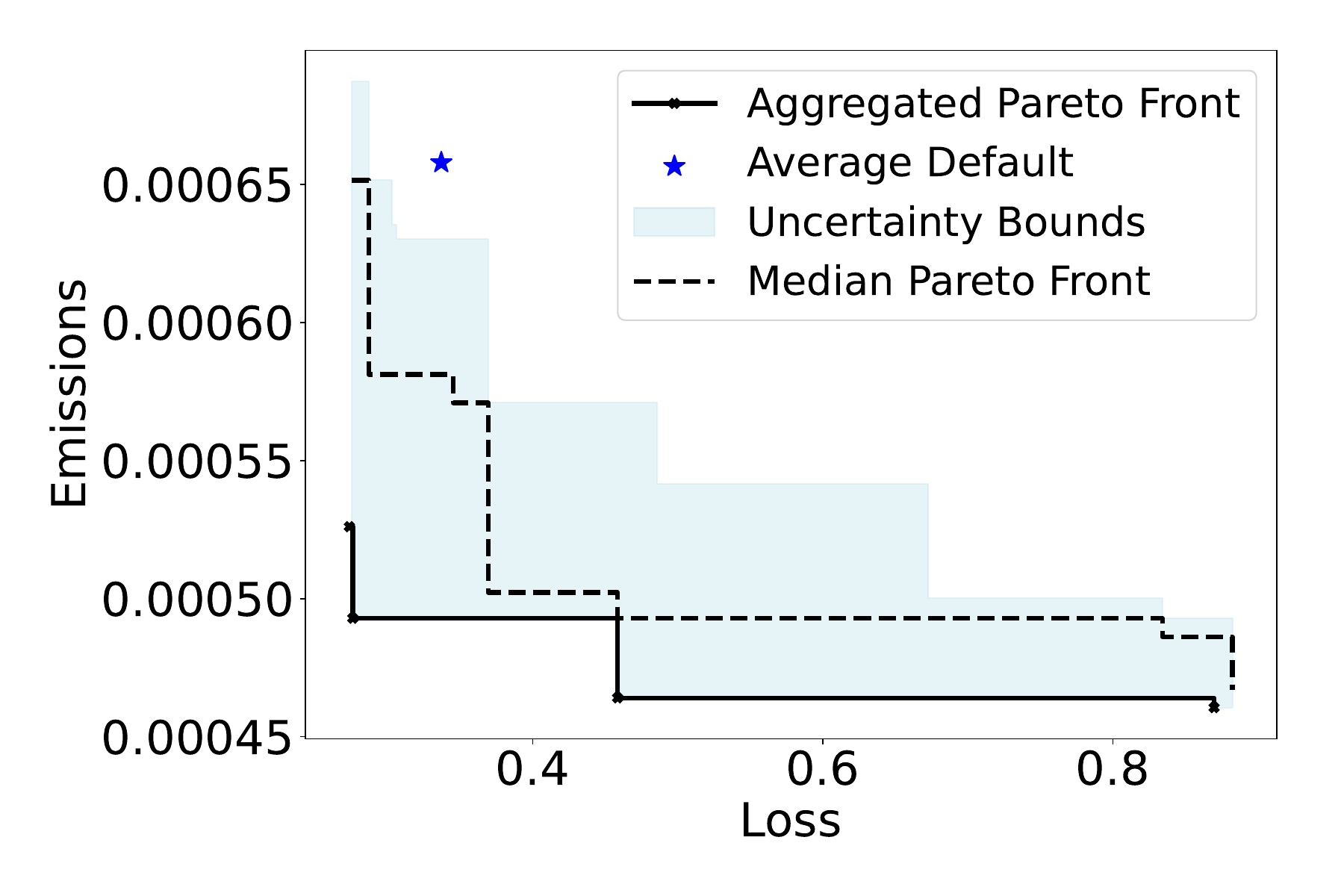}
        \caption{MobileNet on Caltech101}
        \label{fig:mobilenet_caltech}
    \end{subfigure}
    \hfill
    \begin{subfigure}[t]{0.47\textwidth}
        \includegraphics[width=\textwidth]{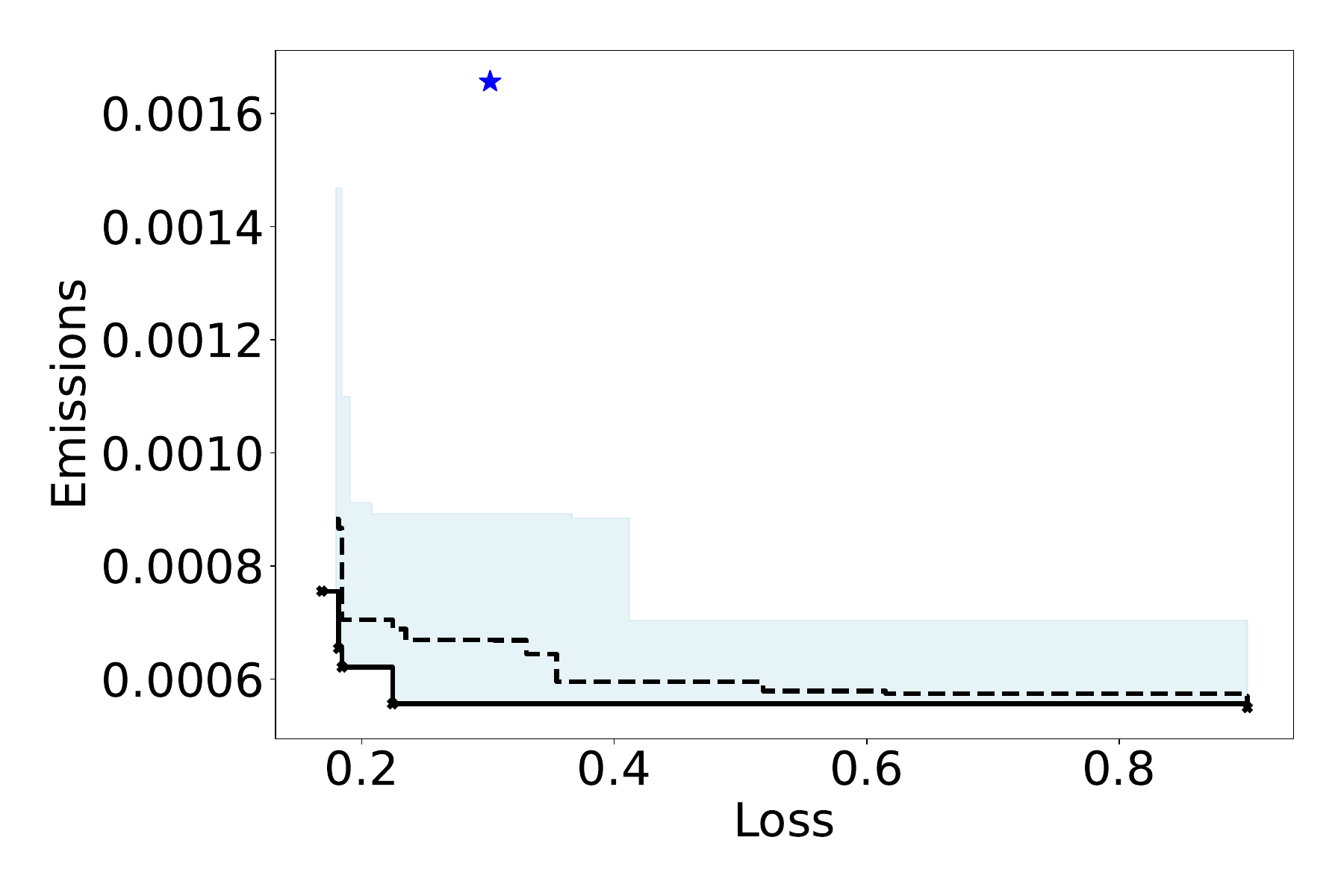}
        \caption{MobileNet on CIFAR10}
        \label{fig:mobilenet_cifar}
    \end{subfigure}

    \vspace{0.5em}

    \begin{subfigure}[t]{0.47\textwidth}
        \includegraphics[width=\textwidth]{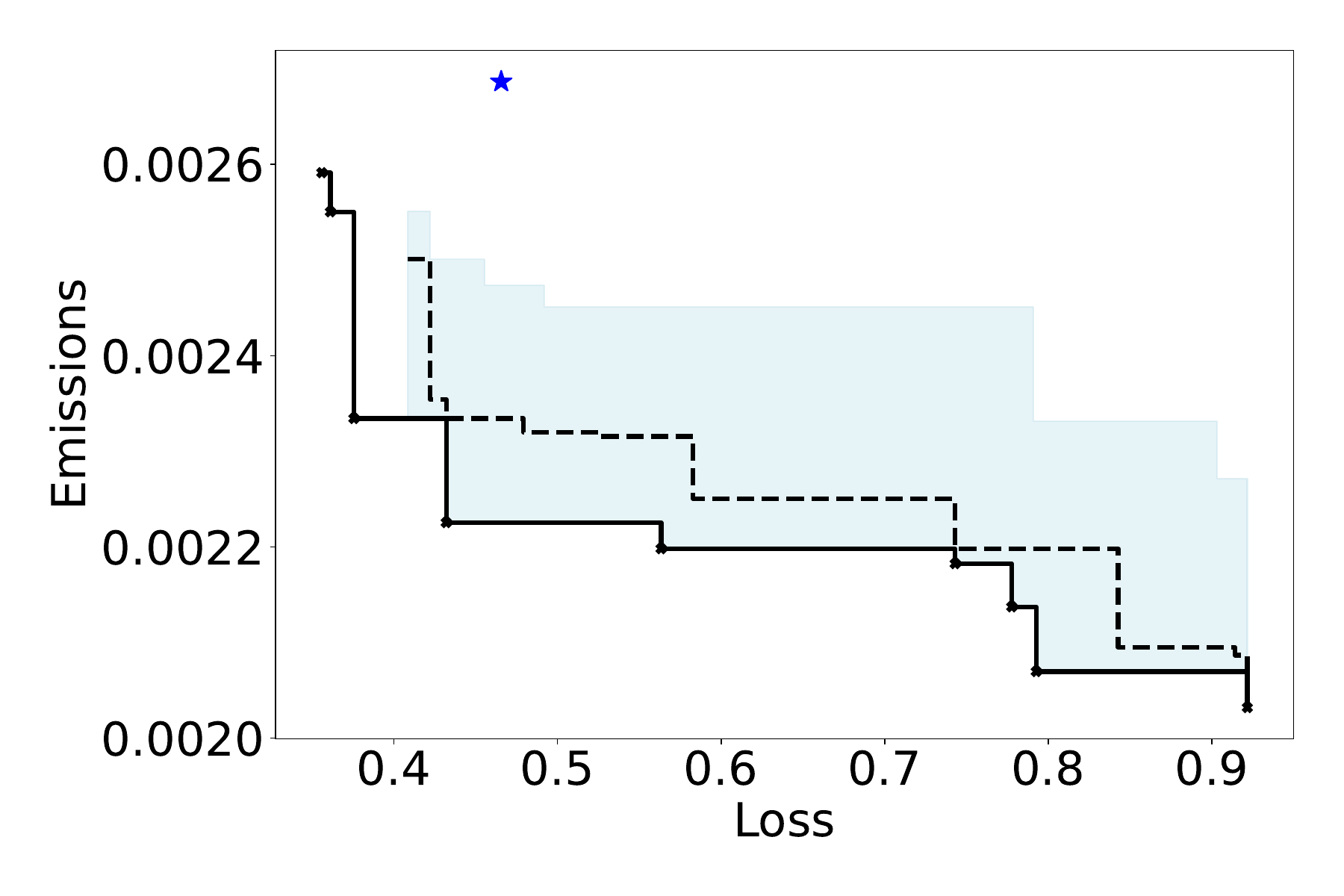}
        \caption{GoogLeNet on Caltech101}
        \label{fig:googlenet_caltech}
    \end{subfigure}
    \hfill
    \begin{subfigure}[t]{0.47\textwidth}
        \includegraphics[width=\textwidth]{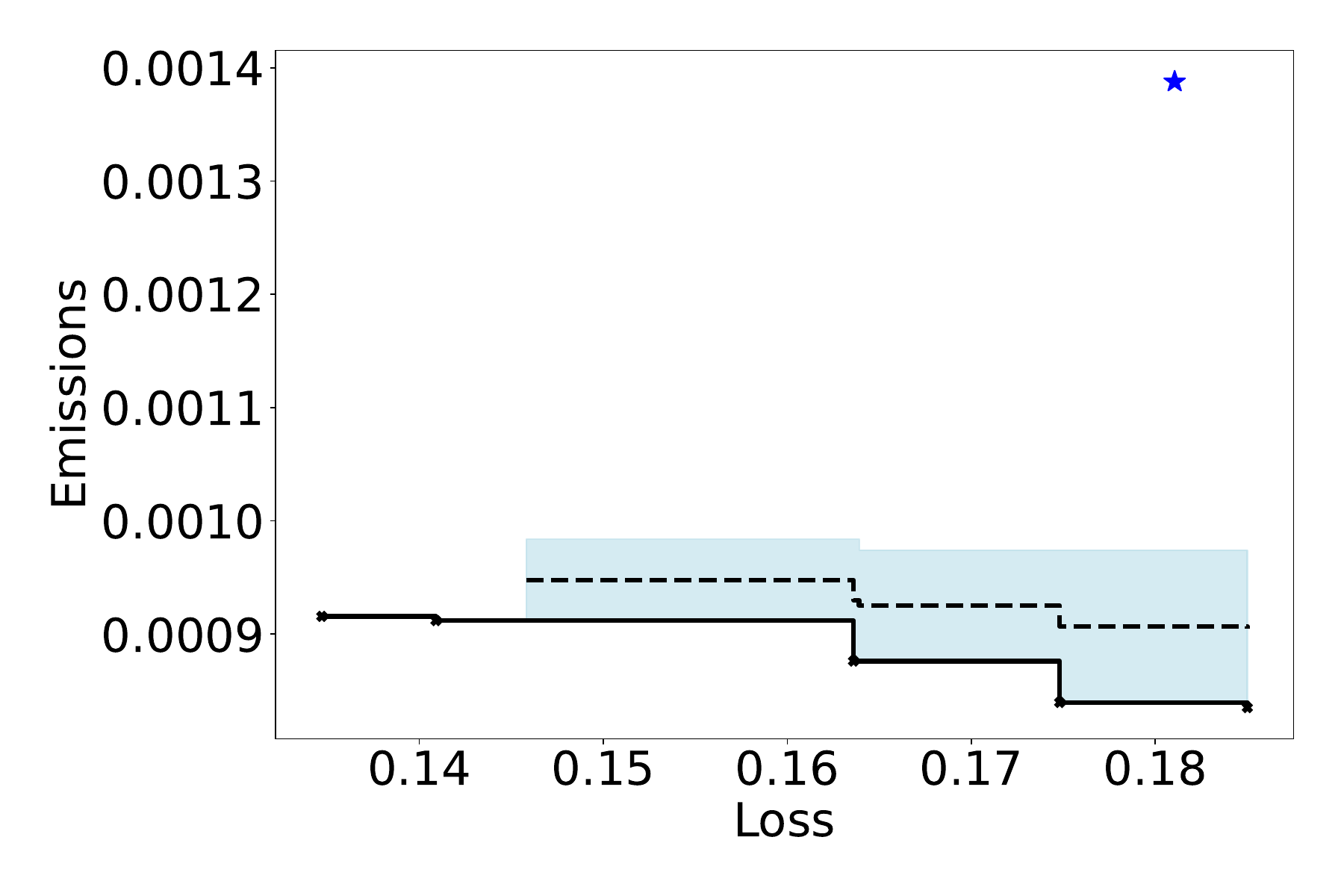}
        \caption{GoogLeNet on CIFAR10}
        \label{fig:googlenet_cifar}
    \end{subfigure}

    \vspace{0.5em}

    \begin{subfigure}[t]{0.47\textwidth}
        \includegraphics[width=\textwidth]{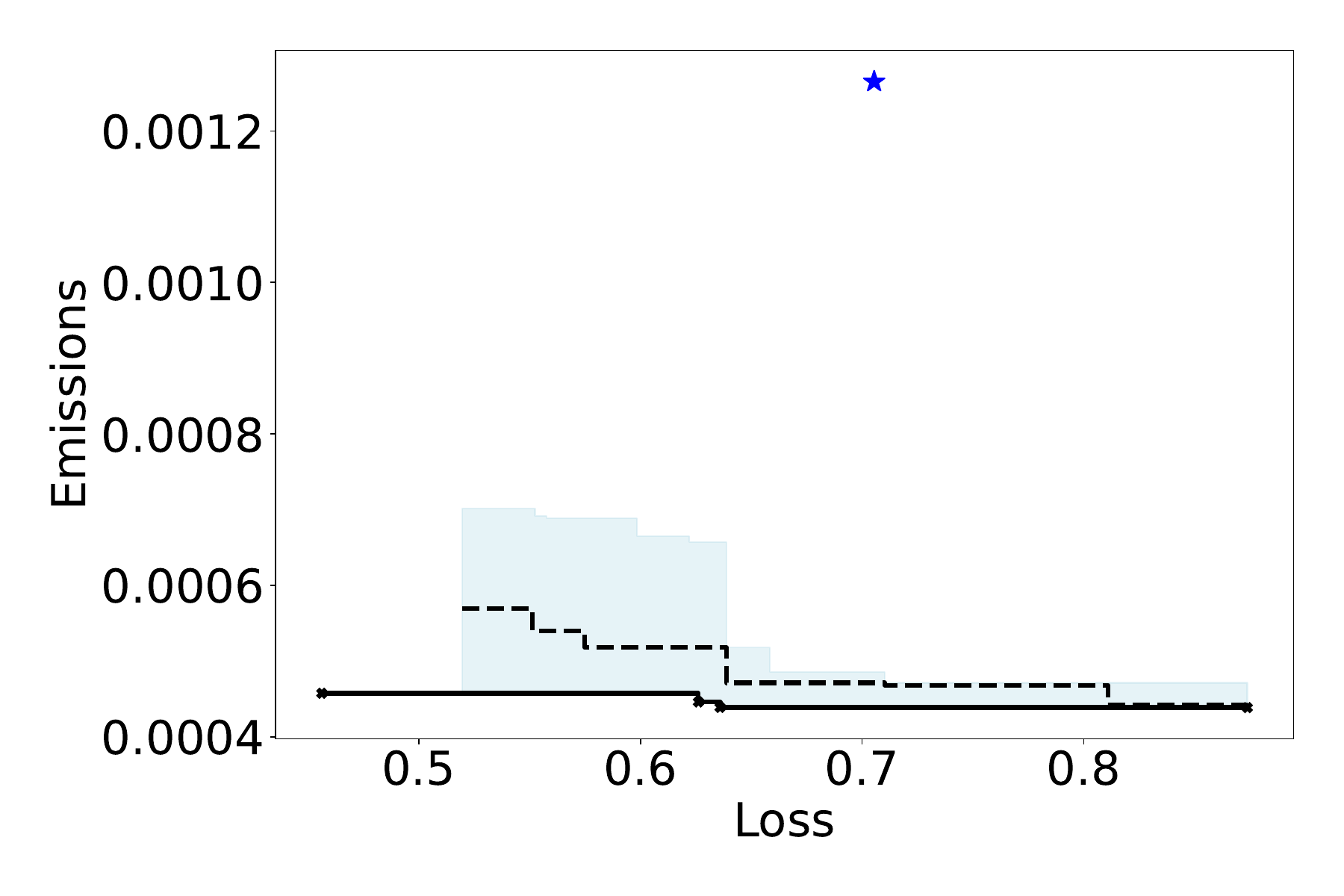}
        \caption{ResNet20 on Caltech101}
        \label{fig:resnet_caltech}
    \end{subfigure}
    \hfill
    \begin{subfigure}[t]{0.47\textwidth}
        \includegraphics[width=\textwidth]{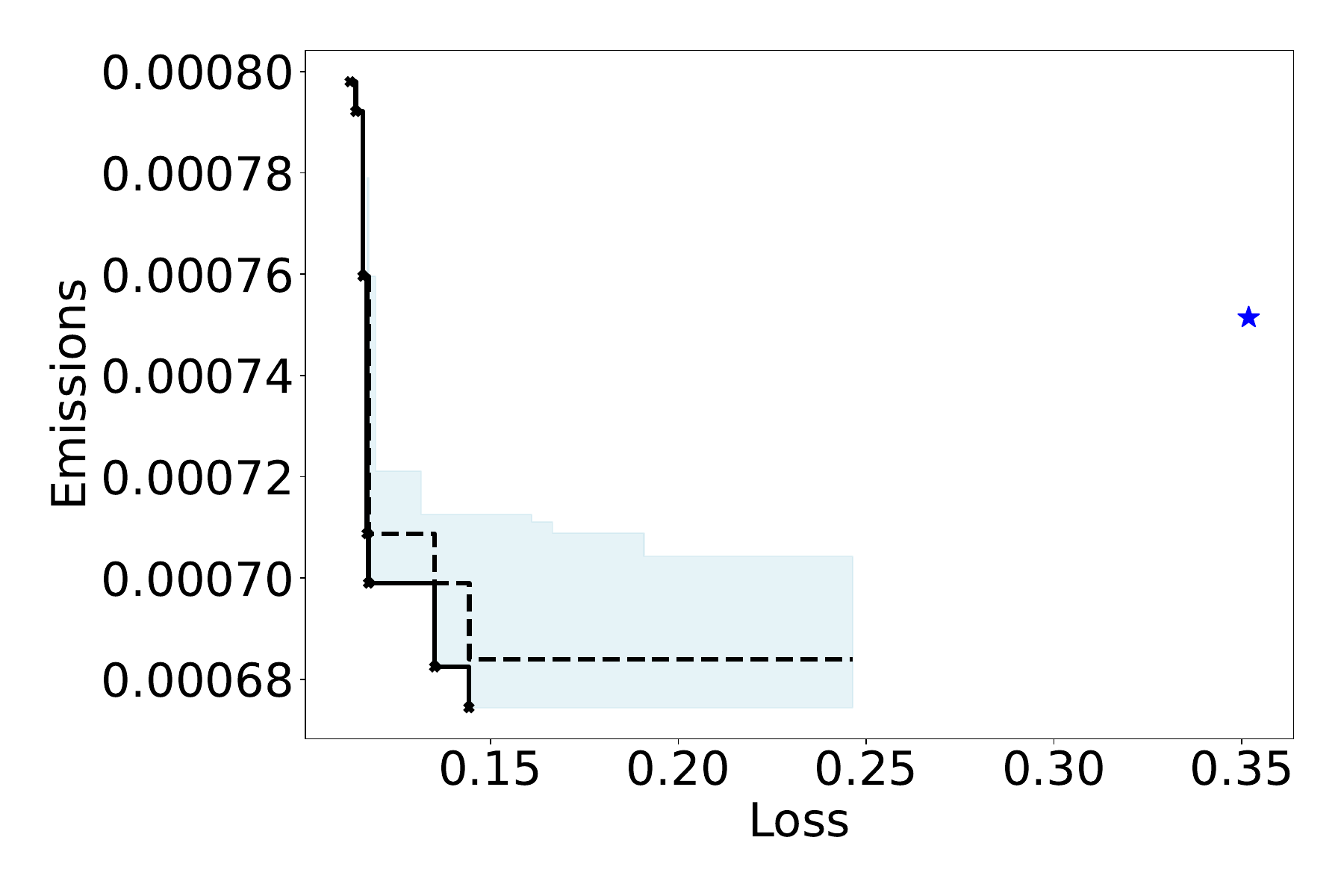}
        \caption{ResNet20 on CIFAR10}
        \label{fig:resnet_cifar}
    \end{subfigure}

    \caption{Comparison of Pareto fronts for MobileNet, GoogLeNet, and ResNet20 on Caltech101 and CIFAR10 datasets over multiple seeds. We show the loss in \% and the emissions in kgCO\textsubscript{2}eq. The plots include median Pareto fronts with uncertainty bars, as well as an aggregated Pareto front of Pareto-optimal solutions across all runs. The star denotes the averaged performance of the default DSNN configuration.}
    \label{fig:all_configs}
\end{figure}

We first discuss the quantitive results,  then the importance of the optimized hyperparameters, and finally, the implications for the configuration space. Note that we focus on the insights gained regarding DSNNs and not on how efficient our (or others') HPO approach is.
In Figure \ref{fig:all_configs}, we present the computed Pareto fronts of a ResNet20, MobileNetV2 and GoogLeNet architecture, optimized with our multi-fidelity multi-objective (MFMO) framework, on the CIFAR10 and Caltech101 datasets.
Shown is a diverse set of optimal configurations that either minimize or balance the primary objectives of model accuracy and energy consumption. These are aggregated results over three seeds. The Pareto front shown results from aggregating the three individual Pareto fronts and extracting the Pareto optimal points. The default value is the mean over the loss and emissions of the default configuration on the three seeds.
The Pareto fronts in Figure \ref{fig:all_configs} depict how each configuration performs relative to the others within the defined hyperparameter space. The configurations were evaluated regarding classification loss and emissions emitted during inference of the model instantiated with the respective configuration. Throughout this paper, when referring to emissions, we measured emissions at inference.
% \captionsetup[figure]{belowskip=1cm, aboveskip=1cm} % Adjust the space above and below the caption
Although we expect that ~\cite{DBLP:conf/cvpr/ElhoushiCSTL21} optimized their hyperparameters at least manually, we can show that our AutoML approach found even better trade-offs of the two objectives.
The default configuration for the DSNN, as designed by~\cite{DBLP:conf/cvpr/ElhoushiCSTL21}, is in fact not part of the Pareto front in Figure \ref{fig:all_configs}. This holds for all architectures on both datasets.
This means that there are better configurations that dominate the default configuration regarding both loss and emissions on MobileNetV2 (Figures~\ref{fig:mobilenet_caltech} and \ref{fig:mobilenet_cifar}), GoogLeNet (Figures \ref{fig:googlenet_caltech} and \ref{fig:googlenet_cifar}) and ResNet20 (Figures~\ref{fig:resnet_caltech} and \ref{fig:resnet_cifar}) on Caltech101 and CIFAR10.

\begin{wrapfigure}{r}{0.7\textwidth} % 'r' means right side, 0.5\textwidth makes it take half the page
    \centering
    \includegraphics[width=\linewidth]{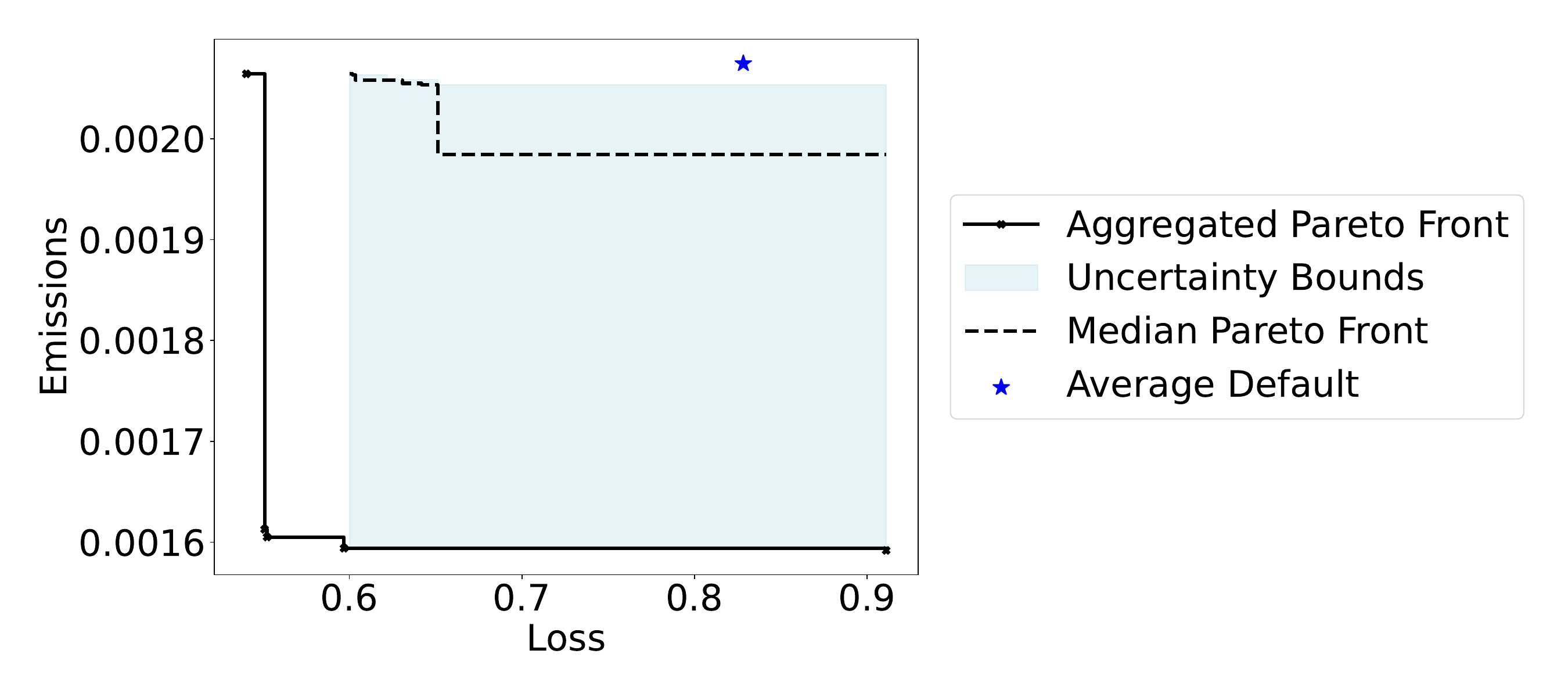}
    \caption{Pareto front for EfficientNetV2 on CIFAR10 over multiple seeds. We show the loss in \% and the emissions in kgCO\textsubscript{2}eq. The plots include median Pareto fronts with uncertainty bars, as well as an aggregated Pareto front of Pareto-optimal solutions across all runs. The star denotes the averaged performance of the default DSNN configuration.}
    \label{efficientnet cifar figure}
\end{wrapfigure}

%For the ResNet20 on CIFAR10 (Figure \ref{fig:pareto_cifar10_resnet20}), the default configurations produce a higher loss but lower emissions than one of the Pareto-dominating solutions.
Having a closer look at Figure \ref{fig:resnet_cifar}, the underlying goal of our MFMO optimization remains to balance performance and efficiency.
Hence, the configurations at the bottom left of the Pareto front are especially relevant since they minimize both objectives simultaneously instead of heavily prioritizing either.
There is an absolute reduction in loss of up to 20\% in these configurations compared to the defaults.
At the same time, relative emission reduction ranges from approx. 10\% to more than 60\% in Figures \ref{fig:mobilenet_cifar} and \ref{fig:resnet_caltech}. Additionally, we achieved a maximum loss reduction of about 20\% for the ResNet20 on CIFAR10.
This validates the need for proper HPO tuning since we found better configurations that take the energy-efficient DSNNs a significant step further by improving their accuracy and energy consumption.

Additionally, we optimized an EfficientNetV2-based DSNN on CIFAR-10 to evaluate whether our method can yield further improvements when applied to an architecture already optimized for efficiency, thereby testing its robustness and generalizability. The results are shown in Figure \ref{efficientnet cifar figure} with the corresponding Pareto-optimal solutions in Table \ref{table efficientnet cifar} in the appendix. Again, we were able to identify multiple configurations that surpass the default DSNN w.r.t. prediciton performance and emissions at inferece, using our MOMF approach. While EfficientNetV2 \citep{DBLP:conf/icml/TanL21} is inherently designed for optmizing for performance and efficiency, using compound scaling, the need for carefully tuning the quantization hyperparameters is apparent. Compared to the default EfficientNetV2, we were able to reduce emissions by more than 20\%.

\subsubsection{Hyperparameter Importances}

Another crucial aspect is the analysis of hyperparameter importance to learn their influence on a model and lay the foundation of our DSNN design insights in the next subsection. We use DeepCAVE~\citep{sass-realml22a} for analyzing the Pareto front of our MFMO analysis in Figure \ref{fig:resnet_cifar}, and computing the hyperparameter importance using fANOVA~\citep{hutter-icml14a}. 

fANOVA fits a random forest surrogate model to the hyperparameter optimization landscape and decomposes the model's variance into components corresponding to hyperparameters. This allows fANOVA to estimate the marginal impact of individual hyperparameters or pairs of hyperparameters. For an extension to MO optimization, fANOVA can be applied to each objective's performance surface separately. The hyperparameter importances are then computed for different weightings of the objectives.

% local parameter importance (LPI)~\citep{biedenkapp-lion18a}. LPI simulates the behavior of developers who are interested in changing only a single hyperparameter at a time and checking its influence on performance. So, LPI assesses the importance of each hyperparameter by observing how small changes to hyperparameters within a localized region of the hyperparameter space affect the model's performance. Specifically, LPI assigns importance to a hyperparameter according to the performance variation along the changes of one hyperparameter. So, LPI is the fraction of performance variance given by a specific ICE curve~\citep{goldstein-jcgs15a} on an incumbent configuration as a reference point.

\begin{figure}[ht]
    \centering
    \begin{subfigure}[t]{\textwidth}
        \centering
        \includegraphics[width=\textwidth]{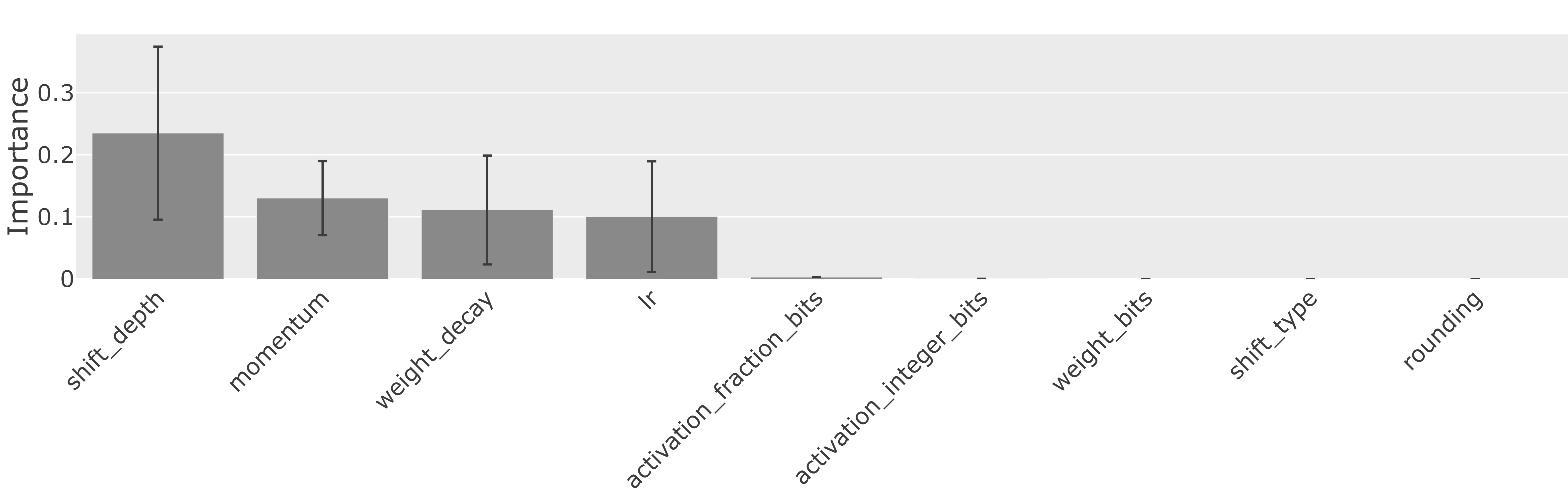}
        \caption{Hyperparameter importance with respect to loss.}
        \label{fig:importances_loss}
    \end{subfigure}
    
    \vspace{0.5cm} % Optional vertical space between the subfigures
    
    \begin{subfigure}[t]{\textwidth}
        \centering
        \includegraphics[width=\textwidth]{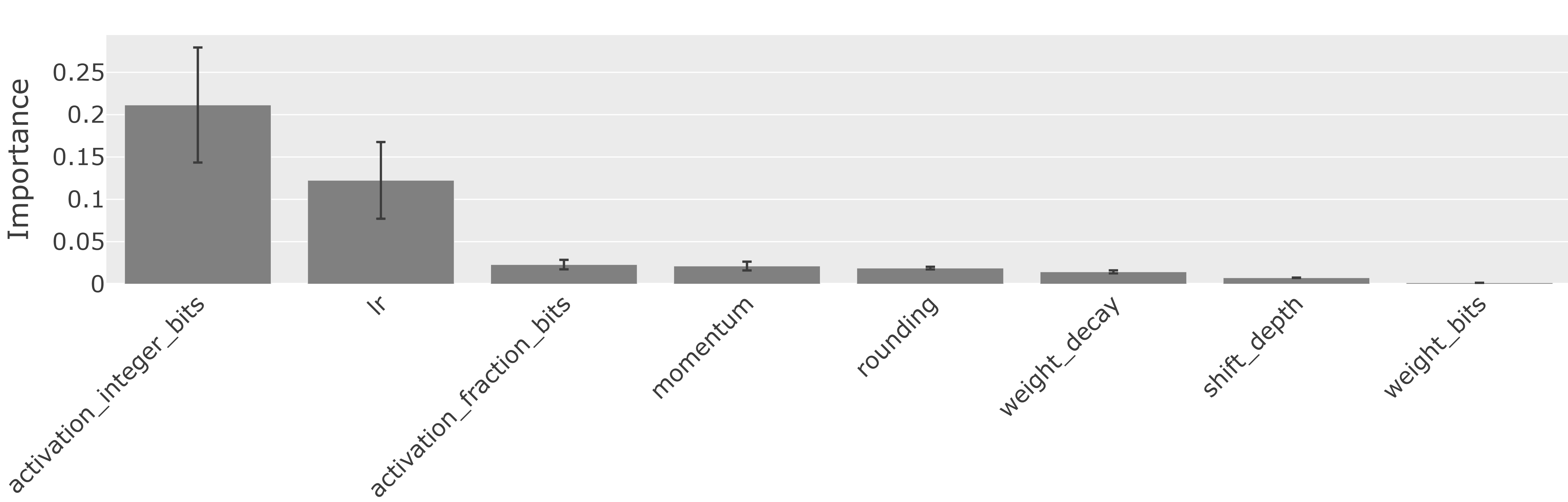}
        \caption{Hyperparameter importance with respect to emissions.}
        \label{fig:importances_emissions}
    \end{subfigure}
    
    \captionsetup{justification=centering}
    \caption{Hyperparameter importance according to fANOVA for ResNet20 on CIFAR10. (a) Importance with respect to loss. (b) Importance with respect to emissions.}
    \label{fig:fanova_importances}
\end{figure}

Figures \ref{fig:importances_loss} and \ref{fig:importances_emissions} show the hyperparameter importances of a ResNet20 on Cifar10 w.r.t. loss and emissions, respectively. The MO-fANOVA analysis for different weightings of the objectives loss and emissions can be found in Figure \ref{fig:resnetcifar_mo_fanova} in the appendix. The most important DSNN-specific hyperparameters for emissions in Figure \ref{fig:importances_emissions} include activation integer and fraction bits. This hints at the precision of the activation quantization being the most controlling factor for energy efficiency. Naturally, precision is a key factor since it controls the amount of operations in the network. Regarding loss in Figure \ref{fig:importances_loss}, the shift depth is the most important hyperparameter. The proportion of the network converted to perform shift operations naturally controls the amount of information retained in the network. This is crucial for the overall performance of the network.

Notably, the shift type has a low importance value in Figure \ref{fig:importances_loss} and is not among the most important hyperparameters in Figure \ref{fig:importances_emissions}.
This could indicate that the shift type is marginally relevant for both objectives.
In practice, this insight could aid in model training by allocating fewer resources to tuning this hyperparameter, allocating resources tailored to the use case. This is further supported by looking at Tables \ref{table resnet cifar} to \ref{table efficientnet cifar}. About 50\% of the Pareto optimal configurations use either shift type, meaning they are not leaning toward either to maximize either objective.

With the analysis of hyperparameter importance in our study, we offer a baseline of hyperparameters to include for future training and inference purposes. Including only the significant ones is a promising way of further boosting the energy efficiency of the DSNNs and the optimization process \citep{probst-jmlr19a}.
Additional plots showing the hyperparameter importance for each architecture can be found in Appendix \ref{appendix importances}. DSNN-specific hyperparameters consistently place among the most influential, alongside weight decay and learning rate. This consistent importance underscores the relevance of understanding and optimizing DSNNs.

\subsubsection{Insights into the Design of DSNNs}
\label{insights subsection}

When looking at the configurations of the ResNet20 architecture on CIFAR10 in Table~\ref{table resnet cifar} (see Appendix for similar tables for the other architectures and datasets), most solutions have a surprisingly small shift depth $s\in\{1,3\}$, compared to 20 as the maximal value and the setting of the default.
% Hence, the Pareto optimal solutions are consistent with a very low shift depth.
At the same time, the number of activation fraction bits is often rather high.
This leads to the assumption that the bulk of information is retained in the fraction part of the activation value.
A valid expectation since we used batch normalization in the ResNet20, same as \cite{DBLP:conf/cvpr/ElhoushiCSTL21}.
In batch normalization, the layer inputs are re-scaled and re-centered using the mean and variance of the corresponding dimension~\citep{ioffe-icml15a}.
This usually leads to small weights, highlighting the importance of activation fraction precision, which is higher in Pareto optimal configurations. It ranges from 8 to 32, likely a contributing factor to the emission reduction.

While we initially hypothesized a direct proportionality between shift depth and emission savings, our analysis reveals a more nuanced relationship between hyperparameters, especially the shift depth and the bit precision in weights and activations.
These hyperparameters interact in a non-linear manner, jointly influencing both energy consumption and model performance in ways that are not immediately intuitive.
The results of our Pareto front analysis indicate that the choice of shift layers, which we expected to correlate positively with performance gains and negatively with loss, does not exhibit such a straightforward relationship.
Instead, our findings highlight the intricate dependencies among architectural design choices, hyperparameter configurations, and their combined impact on energy efficiency. Notably, increasing the number of shift layers does not consistently result in greater emission savings. Weight precision also influences emissions, but no consistent trend was observed in the ResNet20 configurations, indicating the need for dataset-specific tuning.

These insights are corroborated by looking at the additional experiments in the appendix. Additionally to the Resnet20 on CIFAR10, we computed the Pareto fronts of our MOMF analysis on MobileNetV2 and GoogLeNet on CIFAR10. The corresponding results are shown in Figures \ref{fig:mobilenet_cifar} and \ref{fig:googlenet_cifar}. The overall Pareto optimal configurations from multiple seeds can be found Appendix \ref{appendix tables} in Tables \ref{table mobilenet cifar} and \ref{table googlenet cifar}. Again, the shift depths are generally very low, either one or three, with two exceptions of seven and fourteen. The number of activation fraction bits is usually close to the upper bound of 32 bits.

We have also computed Pareto fronts of our MFMO approach on the Caltech101 dataset, for ResNet20, MobileNetV2 and GoogLeNet. The results can be seen in Figures \ref{fig:mobilenet_caltech}, \ref{fig:googlenet_caltech} and \ref{fig:resnet_caltech}. The configurations on the Pareto fronts are detailed in Tables \ref{table resnet caltech}, \ref{table mobilenet caltech} and \ref{googlenet caltech table}. Again, the vast majority of configurations have a low shift depth in the range of $s\in\{1,2,3,4,5,6\}$.
Only two GoogLeNet configurations have a shift depth of eight and 9. This is still relatively low, given that GoogLeNet is a complex architecture with 22 layers in total. As with the previously discussed results, this contributes to the reduction of emissions while not impacting the performance.
Generally, the number of activation fraction and integer bits increases with lower shift depth and vice versa. 
This confirms our previous findings from the thoroughly discussed ResNet20 on CIFAR10. 

\begin{figure} % 'r' means right side, 0.5\textwidth makes it take half the page
    \centering
    \includegraphics[width=0.7\linewidth]{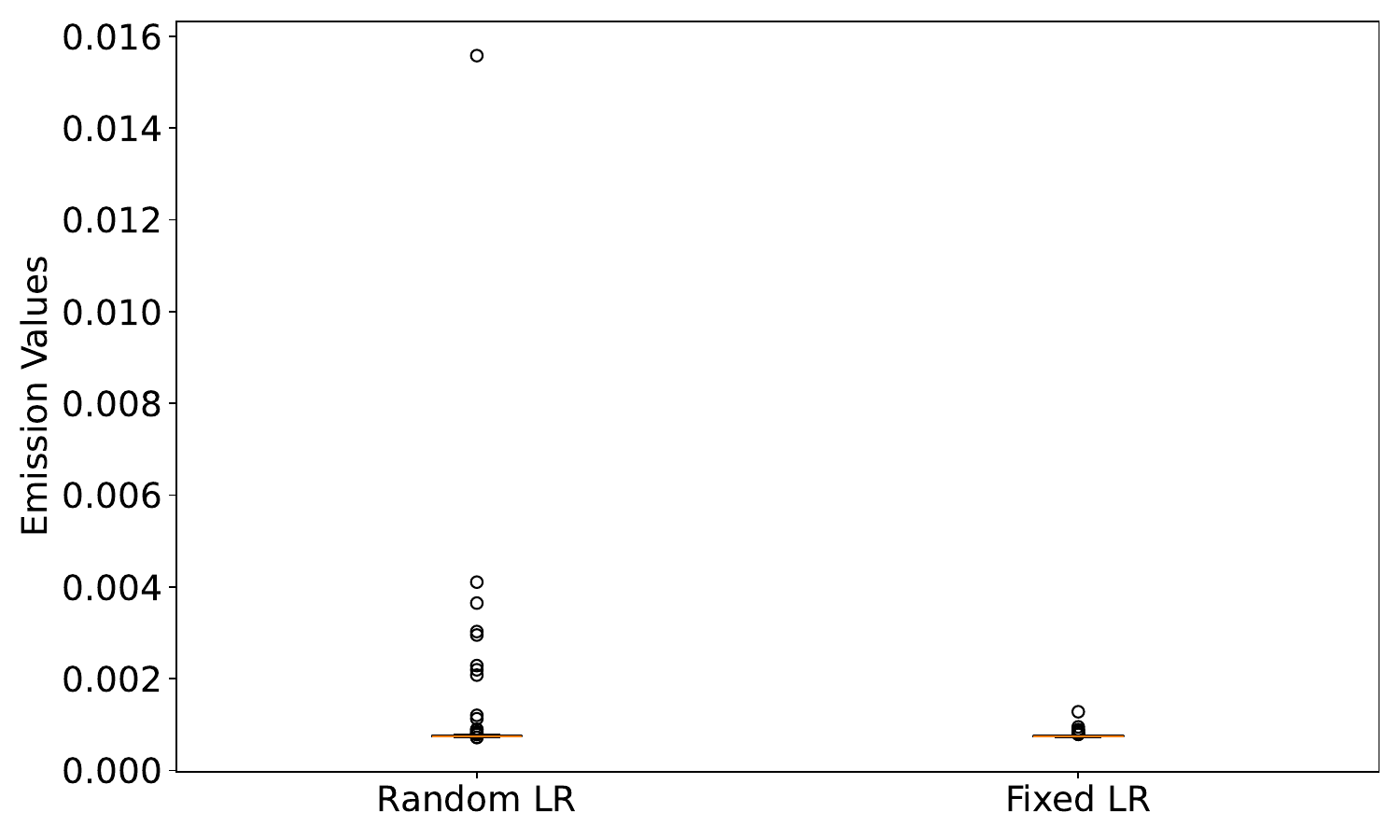}
    \caption{Emissions from different learning rates.}
    \label{fig:emissions_from_lr}
\end{figure}

In Figures \ref{fig:importances_loss} and \ref{fig:importances_emissions}, the learning rate is surprisingly identified as an important hyperparameter both for loss and emissions.
To shed light on how the learning rate is related to emissions, we investigated this further.
To this end, we instantiated a ResNet20 on CIFAR10 with the default configuration, but varied the learning by  100 values randomly sampled from the configuration search space. 
Analyzing the resulting emissions leads to a mean value of 0.00106 kgCO2eq, which exceeds the range of our Pareto fronts.
The standard deviation is 0.00156, which is itself close to the mean value.
This shows, indeed, that randomizing the learning rate introduced great variability in the energy measurements.
As a control measurement to rule out other confounding factors, we measured the energy required for inference for a model instantiated with the default learning rate 100 times over. As expected, this leads to a mean of 0.00076 kgCo2eq, matching the evaluation of the default configuration in Figure \ref{fig:resnet_cifar}, and a very low standard deviation of about 0.00006 (see Figure \ref{fig:emissions_from_lr}).

% 82 of those values are below 0.0008, which is the supremum of our Pareto fronts.
% However, while the majority of the resulting emissions are in the expected range, 13 emission values are significant outliers in a boxplot analysis (Figure \ref{fig:emissions_from_lr}), causing a standard deviation of 0.0009571 kgCo2eq. To ensure that there is a measurable effect, we instantiated a network with the default configuration, using a fixed learning rate of 0.1, and measured the inference emissions XXX times.
% The mean of the resulting emissions is 0.00075 with a standard deviation of XXX.
% This falls within the expected range of below 0.0008, where our Pareto front lies, without outliers.
This leads to the conclusion that the learning rate is indeed an influential HP on the emissions since it can drastically lead to an increase in emissions when chosen suboptimally.
Our AutoML approach offers a solution to exactly this problem, since we aim to automate the identification and optimal choice of hyperparameters.
% \footnote{Note to reviewers: At the time of our submission, we are running the experiments on Caltech101, given in the Appendix, on more seeds to incorporate uncertainty quantification for these in the rebuttal revision of our submission. We have already verified our insights on multiple configurations of DSNNs on two datasets and three architectures; thus we strongly expect consistent results by evaluating more seeds.} 

% Our investigation into Pareto optimal solutions reveals complex trade-offs between performance metrics and emissions and the absence of a linear or direct correlation between the number of shift layers and the model’s energy efficiency and environmental footprint.
% Furthermore, configurations at both extremities of the Pareto front exhibit contradictory outcomes — those with higher emissions do not consistently exhibit lower loss, and vice versa.

\subsection{Transferability of Configurations}

Understanding whether optimal configurations found in one setting remain effective in others is critical for practical and sustainable AutoML.
In particular, the ability to transfer configurations across datasets can lead to substantial reductions in computational cost and associated emissions.
If previously optimized configurations remain near-optimal despite changes in the data, users can avoid restarting the entire optimization process, thus saving both time and resources.
\begin{figure} % 'r' means right side, 0.5\textwidth makes it take half the page
    \centering
    \includegraphics[width=0.7\linewidth]{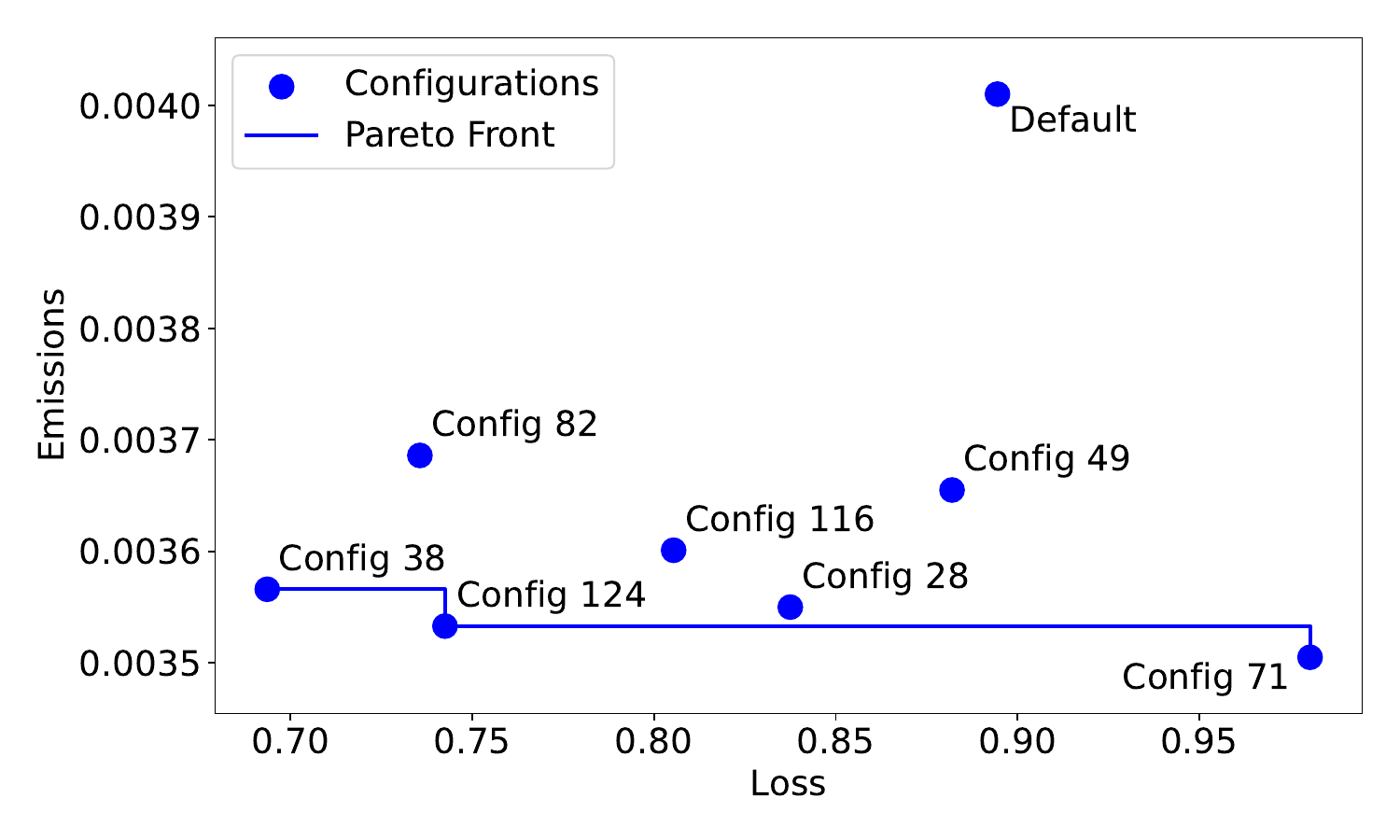}
    \caption{ResNet20 trained on Imagenet100, instantiated with Pareto-optimal configurations from CIFAR10 (see Table \ref{table resnet cifar}). We show the loss in \% and the emissions in kgCO\textsubscript{2}eq.}
    \label{imagenet figure}
\end{figure}
To investigate this, we evaluated the Pareto-optimal configurations identified for a ResNet20 model on CIFAR-10 trained on ImageNet100, a subset of ImageNet \cite{DBLP:conf/eccv/TianKI20}. We measure loss and emissions at inference.
The results, shown in Figure \ref{imagenet figure} demonstrate that several of the original Pareto configurations remain competitive on Imagenet100, continuing to outperform the default configuration. Notably, the default remains Pareto-dominated while the transferred configurations yield improvements in both loss and emissions.
This suggests a degree of robustness and transferability, which could be leveraged in practice to reduce the frequency of full re-optimization runs.

\section{Conclusion}
\label{conclusion}

In this work, we presented our Green AutoML approach towards the sustainable optimization of DSNNs through a multi-fidelity, multi-objective (MFMO) HPO framework. Our approach effectively addressed the critical intersection between advancing the capabilities of Deep Learning and environmental sustainability. By leveraging AutoML tools and integrating the environmental impact as an objective, we adeptly navigated the trade-off between model performance and efficient resource utilization.
Our experimental results focused on a better understanding of DSNNs. We successfully optimized DSNNs to achieve higher accuracy while minimizing energy consumption, surpassing the default configuration settings in both aspects. Through systematic experimentation, we identified key hyperparameters that significantly influence performance and emissions, such as shift depth and number of weight bits.

By optimizing these hyperparameters, our MFMO approach did not just improve one dimension of the problem -- it concurrently enhanced both model loss and energy efficiency, showcasing a balanced improvement across essential performance metrics.
We have thoroughly explored the configuration space for DSNNs, introduced a Green AutoML approach for efficiency-driven model development~\citep{tornede-jair23a}, and provided valuable insights into the design decisions impacting DSNN performance. 

The generalizable insights from our work extend beyond the need to carefully select bitwidths for quantization. Our results reveal that optimal configurations for DSNNs are often counterintuitive and highly dependent on the intricate relations between hyperparameters. For example, we found that low shift depths often achieve superior trade-offs between accuracy and energy efficiency, challenging assumptions about full quantization of networks. Additionally, our analysis highlights the importance of prioritizing specific hyperparameters for different objectives, providing a targeted approach to DSNN optimization. These findings are consistent across multiple architectures and datasets, demonstrating their broader applicability. Overall, our study shows the need for a systematic, automated approach to detecting these insights, which are relevant not only for DSNNs but also for other quantized and resource-efficient neural networks.

\paragraph{Limitations} While our use of CodeCarbon provides valuable insights into the emissions impact of training and inference for DSNNs, it is important to acknowledge the limitations and assumptions inherent in these measurements. CodeCarbon relies on real-time power draw metrics from tools like nvidia-smi. However, these estimates assume a steady power draw during computation and do not account for fluctuations in hardware utilization or dynamic changes in the energy grid.
We consider CodeCarbon a reasonable approximation for measuring energy consumption. This is reflected in recent literature, where studies compared CodeCarbon measurements to wattmeters that directly measure power consumption on machines \citep{bouza2023estimate}.

We acknowledge that energy consumption varies significantly based on regional energy mixes and the infrastructure of data centers. Emissions are typically calculated by multiplying the energy consumed by a carbon intensity factor. For a given energy mix, this factor remains constant. As a result, emissions scale linearly with energy consumption when the energy mix is fixed. Thus, once we compute a Pareto front based on energy consumption, the emission axis can be scaled accordingly to reflect the specific carbon intensity of the region. In dynamic scenarios where energy mixes change over time, our approach enables rapid recalculation of Pareto fronts to reflect these variations so that stakeholders can accommodate real-time or regional changes in energy conditions. While data centers may often rely on carbon-neutral energy sources, the broader applicability of our method extends to embedded devices, such as those used in automobiles or IoT systems. These devices operate in highly varied environments, with energy sources dependent on local conditions. Our approach enables stakeholders to determine location-specific or even time-sensitive Pareto fronts, providing a tailored basis for decision-making. This empowers users to balance trade-offs between accuracy and emissions based on financial, ecological, or operational considerations. 
While testing on hardware such as low-power chips or alternative GPUs could offer additional insights into hardware-specific performance trade-offs, we maximize the potential impact and accessibility of our research by training on NVIDIA A100 GPUs, which is commonly-used hardware in academia and industry.

The DSNN configurations that outperform the default baselines were identified using our MFMO approach. This strategy enabled efficient exploration of a broader region of the search space while keeping training costs manageable. Importantly, the models were not trained to full convergence, by design. Although we expect the accuracy gap between configurations to narrow with full training, our optimal configurations consistently demonstrate higher accuracy in the early stages of training, indicating faster convergence. This approximation — inferring full-training behavior from early-stage performance — is a widely accepted assumption in multi-fidelity optimization. The early stage performance advantage is particularly valuable during optimization, as it enables the rapid identification of promising candidates with significantly reduced computational overhead. Crucially, we also expect the relative emissions savings of these configurations to persist after full training, since the computational graph, operation types, and numerical precision remain fixed throughout training and thus continue to govern energy consumption consistently.

\textcolor{black}{We do not focus on transformer-based architectures in this work.
While transformers have become prominent in computer vision, particularly Vision Transformers (ViTs), they typically require substantial computational resources and are less common in highly constrained environments.
In scenarios such as edge computing and embedded systems for image classification, smaller convolutional architectures remain a predominant choice due to their lower memory footprint, faster inference, and mature optimization toolchains \citep{mauricio2023comparing}.
Although there are emerging quantization strategies for transformers \citep{DBLP:conf/date/SchererCMB24}, our goal is not to replicate or extend the DSNN framework to support these architectures.
In line with the principles of AutoML, we are conscious of the compute budget required for large-scale search and deliberately avoid high-cost models like transformers to ensure that our pipeline aligns with low-resource deployment scenarios.}

Weighing the additional cost of employing AutoML with the savings achieved as a result is vital to assessing the impact of our approach.
We argue that we offset the additional cost of the AutoML training process, namely the generation of the Pareto front for each model, through our savings in inference cost in a negligible amount of time.
We target an application environment like automated driving or automated production, where inference needs to happen at near real-time or real-time.
To provide sufficient evidence, we provide an upper bound calculation based on the experiments we ran. Our longest AutoML optimization run, GoogLeNet on Caltech101, took approximately 48 hours to determine an approximated Pareto front, which is by far the longest; most other runs were completed in about half that time. Assuming a maximum 300W power draw on an A100 GPU according to NVIDIA and a local carbon intensity factor of 0.475 kgCO2/kWh, this results in an estimated emission of about 6.84 kgCO2eq, based on the CodeCarbon formula:
\begin{equation}
    \text{Emissions (kgCO2eq)}
= 0.3\:\text{kW}*48\:\text{h}*0.475\:\text{kgCO2/kWh}
\approx 6.84\:\text{kgCO2eq}.
\end{equation}

For GoogLeNet, our optimized model reduces inference emissions by approximately 0.0004 kgCO2eq per inference, meaning the optimization overhead is amortized after around 17,100 inferences, or about 4.75 hours assuming a conservative real-time inference rate of one image per second. This is much slower than actual rates in automated driving or production contexts, which is what we are targeting. For other models in our study, per-inference emission savings are significantly higher, reaching the break-even point even earlier.

\textcolor{black}{
In this work, we focus exclusively on image classification tasks. Other vision tasks, such as object detection, semantic segmentation, or visual tracking, are relevant for energy-efficient computing in real-world applications, particularly in our intended domains like automated driving and industrial monitoring. However, extending our methodology to these more complex tasks is beyond the scope of this study.}

\paragraph{Future Work} could focus on revisiting the multi-fidelity-multi-objective implementation to find a more efficient way for ParEGO and HyperBand to intertwine, such as by finding a more effective way to assign budgets and weights of objectives. %It is a part of our efforts to lighten the computational load when computing the MFMO Pareto fronts.
Further investigations could include exploring more DSNN-specific fidelity types and multi-objective algorithms to achieve even greater reductions in model emissions. We consider it especially interesting to use the number of weight bits as a fidelity type. By controlling the precision of the weight quantization, training can be sped up in the earlier fidelity while regaining as much information as possible, to use this for full training of the most promising configurations at maximum precision.
Through these future initiatives, the underlying optimization methodology and the environmental benefits of our optimized DSNNs can be further improved, thereby contributing significantly to sustainable AI.

\section*{Acknowledgements}
% By using the \texttt{ack} environment to insert your (optional) 
% acknowledgements, you can ensure that the text is suppressed whenever 
% you use the \texttt{doubleblind} option. In the final version, 
% acknowledgements may be included on the extra page intended for references.
The authors were supported by the German Federal Ministry
for the Environment, Climate Action, Nature Conservation and Nuclear Safety (BMUKN) (GreenAutoML4FAS project no. 67KI32007A).
The authors gratefully acknowledge the computing time provided to them on the high-performance computers Noctua2 at the NHR Center PC2 under the project hpc-prf-intexml. These are funded by the Federal Ministry of Education and Research and the state governments participating on the basis of the resolutions of the GWK for the national high performance computing at universities (www.nhr-verein.de/unsere-partner).

%%%%%%%%%%%%%%%%%%%%%%%%%%%%%%%%%%%%%%%%%%%%%%%%%%%%%%%%%%%%%%%%%%%%%%%%
\clearpage
\bibliographystyle{tmlr}
\bibliography{strings, lib, tmlr, proc}

@article{bischl-dmkd23a,
  title        = {Hyperparameter Optimization: Foundations, Algorithms, Best Practices, and Open Challenges},
  author       = {B. Bischl and M. Binder and M. Lang and T. Pielok and J. Richter and S. Coors and J. Thomas and T. Ullmann and M. Becker and A.{-}L. Boulesteix and D. Deng and M. Lindauer},
  year         = 2023,
  journal      = wileyirdmkd,
  publisher    = {Wiley Online Library},
  pages        = {e1484},
}

@article{chen-arxiv20a,
  title        = {Big self-supervised models are strong semi-supervised learners},
  author       = {T. Chen and S. Kornblith and K. Swersky and M. Norouzi and G. Hinton},
  year         = 2020,
  journal      = {arXiv:2006.10029v2 [cs.LG]},
}

@inproceedings{courbariaux-nips15a,
  title        = {Binaryconnect: Training deep neural networks with binary weights during propagations},
  author       = {M. Courbariaux and Y. Bengio and J. David},
  volume       = 28,
  crossref     = {nips15},
}

@inproceedings{eggensperger-neuripsdbt21a,
  title        = {{HPOBench}: A Collection of Reproducible Multi-Fidelity Benchmark Problems for {HPO}},
  author       = {K. Eggensperger and P. M{\"u}ller and N. Mallik and M. Feurer and R. Sass and A. Klein and N. Awad and M. Lindauer and F. Hutter},
  crossref     = {neuripsdbt21},
}

@article{elsken-jmlr19a,
  title        = {Neural {A}rchitecture {S}earch: A Survey},
  author       = {T. Elsken and J. Metzen and F. Hutter},
  year         = 2019,
  journal      = jmlr,
  volume       = 20,
  number       = 55,
  pages        = {1--21},
}

@article{frazier-jc09a,
  title        = {The Knowledge-Gradient Policy for Correlated Normal Beliefs},
  author       = {P. Frazier and W. Powell and S. Dayanik},
  year         = 2009,
  journal      = {Journal on Computing},
  publisher    = {{INFORMS}},
  volume       = 21,
  number       = 4,
  pages        = {599--613},
}

@article{hennig-jmlr12a,
  title        = {Entropy search for information-efficient global optimization},
  author       = {P. Hennig and C. Schuler},
  year         = 2012,
  journal      = jmlr,
  volume       = 98888,
  number       = 1,
  pages        = {1809--1837},
}

@inproceedings{hernandez-nips14a,
  title        = {Predictive Entropy Search for Efficient Global Optimization of Black-box Functions},
  author       = {J. Hern\'{a}ndez-Lobato and M. Hoffman and Z. Ghahramani},
  crossref     = {nips14},
}

@inproceedings{hutter-icml14a,
  title        = {An Efficient Approach for Assessing Hyperparameter Importance},
  author       = {F. Hutter and H. Hoos and K. Leyton-Brown},
  pages        = {754--762},
  crossref     = {icml14},
}

@inproceedings{hutter-lion11a,
  title        = {Sequential Model-Based Optimization for General Algorithm Configuration},
  author       = {F. Hutter and H. Hoos and K. Leyton-Brown},
  pages        = {507--523},
  crossref     = {lion11},
}

@inproceedings{ioffe-icml15a,
  title        = {Batch Normalization: Accelerating Deep Network Training by Reducing Internal Covariate Shift},
  author       = {Ioffe, S. and Szegedy, C.},
  year         = 2015,
  crossref     = {icml15},
}

@inproceedings{jamieson-aistats16a,
  title        = {Non-stochastic Best Arm Identification and {H}yperparameter {O}ptimization},
  author       = {K. Jamieson and A. Talwalkar},
  year         = 2016,
  crossref     = {aistats16},
}

@article{lindauer-jmlr22a,
  title        = {{SMAC3}: A Versatile Bayesian Optimization Package for {H}yperparameter {O}ptimization},
  author       = {M. Lindauer and K. Eggensperger and M. Feurer and A. Biedenkapp and D. Deng and C. Benjamins and T. Ruhkopf and R. Sass and F. Hutter},
  year         = 2022,
  journal      = jmlr,
  volume       = 23,
  number       = 54,
  pages        = {1--9},
}

@article{moraleshernandez-air22a,
  author = {A. Morales-Hernández and I. Van Nieuwenhuyse and S. Gonzalez},
  title = {A survey on multi-objective hyperparameter optimization algorithms for Machine Learning},
  year = {2022},
  journal = {Artificial Intelligence Review},   
  volume={56},
  pages={8043-–8093}
}

@article{probst-jmlr19a,
  title        = {Tunability: Importance of Hyperparameters of Machine Learning Algorithms},
  author       = {P. Probst and A. Boulesteix and B. Bischl},
  year         = 2019,
  journal      = jmlr,
  volume       = 20,
  number       = 53,
  pages        = {1--32},
}

@inproceedings{sandler-cvpr18a,
  title        = {MobileNetV2: Inverted Residuals and Linear Bottlenecks},
  author       = {M. Sandler and A. Howard and M. Zhu and A. Zhmoginov and L. Chen},
  crossref     = {cvpr18},
}

@inproceedings{sass-realml22a,
  title        = {DeepCAVE: An Interactive Analysis Tool for Automated Machine Learning},
  author       = {R. Sass and E. Bergman and A. Biedenkapp and F. Hutter and M. Lindauer},
  year         = 2022,
  crossref     = {realml22},
}

@inproceedings{schmucker-metalearn20a,
    author = {R. Schmucker and M. Donini and V. Perrone and M. Zafar and C. Archambeau},
    title = {Multi-Objective Multi-Fidelity Hyperparameter Optimization with Application to Fairness},
    crossref = {metalearn20}
}

@article{shahriari-ieee16a,
  title        = {Taking the Human Out of the Loop: {A} Review of {B}ayesian Optimization},
  author       = {B. Shahriari and K. Swersky and Z. Wang and R. Adams and N. de Freitas},
  year         = 2016,
  journal      = {Proceedings of the {IEEE}},
  volume       = 104,
  number       = 1,
  pages        = {148--175},
}

@article{simonyan-arxiv14a,
  title        = {Very Deep Convolutional Networks for Large-Scale Image Recognition},
  author       = {K. Simonyan and A. Zisserman},
  year         = 2014,
  journal      = {arXiv:1409.1556 [cs.CV]},
}

@article{tornede-jair23a,
  title        = {Towards Green Automated Machine Learning: Status Quo and Future Directions},
  author       = {T. Tornede and A. Tornede and J. Hanselle and F. Mohr and M. Wever and E. H{\"{u}}llermeier},
  year         = 2023,
  journal      = jair,
  volume       = 77,
  pages        = {427--457},
}

@proceedings{aistats16,
  title        = {Proceedings of the Seventeenth International Conference on Artificial Intelligence and Statistics ({AISTATS}'16)},
  year         = 2016,
  booktitle    = {Proceedings of the Seventeenth International Conference on Artificial Intelligence and Statistics ({AISTATS}'16)},
  publisher    = {Proceedings of Machine Learning Research},
  volume       = 51,
  editor       = {A. Gretton and C. Robert},
}

@proceedings{cvpr18,
  title        = {Proceedings of the International Conference on Computer Vision and Pattern Recognition ({CVPR}'18)},
  year         = 2018,
  booktitle    = {Proceedings of the International Conference on Computer Vision and Pattern Recognition ({CVPR}'18)},
  publisher    = ieee,
  organization = cvfandieee,
}

@proceedings{icml14,
  title        = {Proceedings of the 31th International Conference on Machine Learning, ({ICML}'14)},
  year         = 2014,
  booktitle    = {Proceedings of the 31th International Conference on Machine Learning, ({ICML}'14)},
  publisher    = {Omnipress},
  editor       = {E. Xing and T. Jebara},
}

@proceedings{icml15,
  title        = {Proceedings of the 32nd International Conference on Machine Learning ({ICML}'15)},
  year         = 2015,
  booktitle    = {Proceedings of the 32nd International Conference on Machine Learning ({ICML}'15)},
  publisher    = {Omnipress},
  volume       = 37,
  editor       = {F. Bach and D. Blei},
}

@proceedings{lion11,
  title        = {Proceedings of the Fifth International Conference on Learning and Intelligent Optimization ({LION}'11)},
  year         = 2011,
  booktitle    = {Proceedings of the Fifth International Conference on Learning and Intelligent Optimization ({LION}'11)},
  publisher    = springer,
  series       = lncs,
  volume       = 6683,
  editor       = {C. Coello},
}

@proceedings{metalearn20,
  title        = {Neur{IP}S 2020 Workshop on Meta-Learning},
  year         = 2020,
  booktitle    = {Neur{IP}S 2020 Workshop on Meta-Learning},
  editor       = {R. Calandra and J. Clune and E. Grant and J. Schwarz and J. Vanschoren and F. Visin and J. Wang},
}

@proceedings{neuripsdbt21,
  title        = {Proceedings of the Neural Information Processing Systems Track on Datasets and Benchmarks},
  year         = 2021,
  booktitle    = {Proceedings of the Neural Information Processing Systems Track on Datasets and Benchmarks},
  publisher    = curran,
  editor       = {J. Vanschoren and S. Yeung},
}

@proceedings{nips14,
  title        = {Proceedings of the 28th International Conference on Advances in Neural Information Processing Systems ({N}eur{IPS}'14)},
  year         = 2014,
  booktitle    = {Proceedings of the 28th International Conference on Advances in Neural Information Processing Systems ({N}eur{IPS}'14)},
  publisher    = curran,
  editor       = {Z. Ghahramani and M. Welling and C. Cortes and N. Lawrence and K. Weinberger},
}

@proceedings{nips15,
  title        = {Proceedings of the 29th International Conference on Advances in Neural Information Processing Systems ({N}eur{IPS}'15)},
  year         = 2015,
  booktitle    = {Proceedings of the 29th International Conference on Advances in Neural Information Processing Systems ({N}eur{IPS}'15)},
  publisher    = curran,
  editor       = {C. Cortes and N. Lawrence and D. Lee and M. Sugiyama and R. Garnett},
}

@proceedings{realml22,
  title        = {{ICML} Workshop on Adaptive Experimental Design and Active Learning in the Real World (ReALML Workshop 2022)},
  year         = 2022,
  booktitle    = {{ICML} Adaptive Experimental Design and Active Learning in the Real World (ReALML Workshop 2022)},
  editor       = {M. Mutny and I. Bogunovic and W. Neiswanger and S. Ermon and Y. Yue and A. Krause},
}

@STRING{acm     = "ACM Press" }

@STRING{aaai    = "Proceedings of the National Conference on Artificial
                  Intelligence (AAAI)" }

@STRING{ai      = "Artificial Intelligence" }

@STRING{curran  = "Curran Associates" }

@STRING{ieee = "IEEE" }

@STRING{jair    = "Journal of Artificial Intelligence Research" }

@STRING{jmlr    = "Journal of Machine Learning Research" }

@STRING{lncs    = "Lecture Notes in Computer Science" }

@STRING{springer = "Springer" }

@STRING{wiley   = "John Wiley \& sons" }

@STRING{cvfandieee     = "Computer Vision Foundation and IEEE Computer Society"}

@STRING{wileyirdmkd   = "Wiley Interdisciplinary Reviews: Data Mining and Knowledge Discovery"}

@article{DBLP:journals/jgo/JonesSW98,
	title = {Efficient Global Optimization of Expensive Black-box Functions},
	author = {D. Jones and M. Schonlau and W. Welch},
	journal = {J. Glob. Optim.},
	volume = {13},
	number = {4},
	pages = {455--492},
	year = {1998}
}

@book{DBLP:books/lib/RasmussenW06,
	title = {Gaussian Processes for Machine Learning},
	author = {C. Edward Rasmussen and C. Williams},
	year = {2006}
}

@inproceedings{DBLP:conf/cvpr/ElhoushiCSTL21,
	title = {{DeepShift}: Towards Multiplication-less Neural Networks},
	author = {M. Elhoushi and Z. Chen and F. Shafiq and Y. Henry Tian and J. Yiwei Li},
	booktitle = {{IEEE} Conference on Computer Vision and Pattern Recognition Workshops,
                  {CVPR} Workshops 2021, virtual, June 19-25, 2021},
	pages = {2359--2368},
	year = {2021}
}

@book{DBLP:books/sp/HKV2019,
	title = {Automated Machine Learning - Methods, Systems, Challenges},
	editor = {F. Hutter and L. Kotthoff and J. Vanschoren},
	year = {2019}
}

@article{DBLP:journals/jmlr/LiJDRT17,
	title = {Hyperband: {A} Novel Bandit-based Approach to Hyperparameter Optimization},
	author = {L. Li and K. Jamieson and G. DeSalvo and A. Rostamizadeh and A. Talwalkar},
	journal = {J. Mach. Learn. Res.},
	volume = {18},
	pages = {185:1--185:52},
	year = {2017}
}

@article{DBLP:journals/tec/Knowles06,
	title = {{ParEGO}: a Hybrid Algorithm With On-line Landscape Approximation for Expensive Multiobjective Optimization Problems},
	author = {J. Knowles},
	journal = {{IEEE} Trans. Evol. Comput.},
	volume = {10},
	number = {1},
	pages = {50--66},
	year = {2006}
}

@article{DBLP:journals/network/LiOD18,
	title = {Learning {IoT} in Edge: Deep Learning for the Internet of Things With Edge Computing},
	author = {H. Li and K. Ota and M. Dong},
	journal = {{IEEE} Netw.},
	volume = {32},
	number = {1},
	pages = {96--101},
	year = {2018}
}

@article{krizhevsky2009learning,
	title = {Learning Multiple Layers of Features From Tiny Images},
	author = {A. Krizhevsky and G. Hinton and o.},
	journal = {University of Toronto},
	year = {2009}
}

@article{DBLP:journals/corr/abs-1911-08354,
	title = {Energy Usage Reports: Environmental Awareness as Part of Algorithmic Accountability},
	author = {K. Lottick and S. Susai and S. Friedler and J. Wilson},
	journal = {arXiv:1911.08354},
	year = {2019}
}

@article{DBLP:journals/corr/abs-1910-09700,
	title = {Quantifying the Carbon Emissions of Machine Learning},
	author = {A. Lacoste and A. Luccioni and V. Schmidt and T. Dandres},
	journal = {arXiv:1910.09700},
	year = {2019}
}

@inproceedings{DBLP:conf/cvpr/HeZRS16,
	title = {Deep Residual Learning for Image Recognition},
	author = {K. He and X. Zhang and S. Ren and J. Sun},
	booktitle = {2016 {IEEE} Conference on Computer Vision and Pattern Recognition,
                  {CVPR} 2016, Las Vegas, NV, USA, June 27-30, 2016},
	pages = {770--778},
	year = {2016}
}

@inbook{Deb2014,
	title = {Multi-objective Optimization},
	author = {K. Deb},
	pages = {403--449},
	year = {2014}
}

@article{DBLP:journals/corr/abs-1905-02370,
	title = {Bayesian Optimization for Multi-objective Optimization and Multi-point Search},
	author = {T. Wada and H. Hino},
	journal = {arXiv:1905.02370},
	year = {2019}
}

@article{DBLP:journals/corr/HowardZCKWWAA17,
	title = {{MobileNets}: Efficient Convolutional Neural Networks for Mobile Vision Applications},
	author = {A. Howard and M. Zhu and B. Chen and D. Kalenichenko and W. Wang and T. Weyand and M. Andreetto and H. Adam},
	journal = {arXiv:1704.04861},
	year = {2017}
}

@article{DBLP:journals/pieee/ZhouCLZLZ19,
	title = {Edge Intelligence: Paving the Last Mile of Artificial Intelligence With Edge Computing},
	author = {Z. Zhou and X. Chen and E. Li and L. Zeng and K. Luo and J. Zhang},
	journal = {Proc. {IEEE}},
	volume = {107},
	number = {8},
	pages = {1738--1762},
	year = {2019}
}

@article{DBLP:journals/pieee/SzeCYE17,
	title = {Efficient Processing of Deep Neural Networks: {A} Tutorial and Survey},
	author = {V. Sze and Y. Chen and T. Yang and J. Emer},
	journal = {Proc. {IEEE}},
	volume = {105},
	number = {12},
	pages = {2295--2329},
	year = {2017}
}

@article{DBLP:journals/cacm/SchwartzDSE20,
	title = {Green {AI}},
	author = {R. Schwartz and J. Dodge and N. Smith and O. Etzioni},
	journal = {Commun. {ACM}},
	volume = {63},
	number = {12},
	pages = {54--63},
	year = {2020}
}

@article{DBLP:journals/jair/KandasamyDOSP19,
	title = {Multi-fidelity Gaussian Process Bandit Optimisation},
	author = {K. Kandasamy and G. Dasarathy and J. Oliva and J. Schneider and B. P{\'{o}}czos},
	journal = {J. Artif. Intell. Res.},
	volume = {66},
	pages = {151--196},
	year = {2019}
}

@article{Belakaria2020-re,
	title = {{Multi-Fidelity} {Multi-Objective} Bayesian Optimization: An Output Space Entropy Search Approach},
	author = {S. Belakaria and A. Deshwal and J. Doppa},
	journal = {AAAI},
	volume = {34},
	number = {06},
	pages = {10035--10043},
	year = {2020}
}

@article{DBLP:journals/corr/RastegariORF16,
	title = {Xnor-net: {ImageNet} Classification Using Binary Convolutional Neural Networks},
	author = {M. Rastegari and V. Ordonez and J. Redmon and A. Farhadi},
	journal = {arXiv:1603.05279},
	year = {2016}
}

@inproceedings{DBLP:conf/cvpr/LiFP04,
	title = {Learning Generative Visual Models From Few Training Examples: An Incremental Bayesian Approach Tested on 101 Object Categories},
	author = {L. Fei{-}Fei and R. Fergus and P. Perona},
	booktitle = {{IEEE} Conference on Computer Vision and Pattern Recognition Workshops,
                  {CVPR} Workshops 2004, Washington, DC, USA, June 27 - July 2, 2004},
	pages = {178},
	year = {2004}
}

@article{DBLP:journals/corr/abs-2310-11453,
	title = {{BitNet}: Scaling 1-bit Transformers for Large Language Models},
	author = {H. Wang and S. Ma and L. Dong and S. Huang and H. Wang and L. Ma and F. Yang and R. Wang and Y. Wu and F. Wei},
	journal = {arXiv:2310.11453},
	year = {2023}
}

@inproceedings{DBLP:conf/cvpr/SzegedyLJSRAEVR15,
	title = {Going Deeper With Convolutions},
	author = {C. Szegedy and W. Liu and Y. Jia and P. Sermanet and S. Reed and D. Anguelov and D. Erhan and V. Vanhoucke and A. Rabinovich},
	booktitle = {{IEEE} Conference on Computer Vision and Pattern Recognition, {CVPR}
                  2015, Boston, MA, USA, June 7-12, 2015},
	pages = {1--9},
	year = {2015}
}

@inproceedings{DBLP:conf/cvpr/ChenWXSX0X20,
	title = {{AdderNet}: Do We Really Need Multiplications in Deep Learning?},
	author = {H. Chen and Y. Wang and C. Xu and B. Shi and C. Xu and Q. Tian and C. Xu},
	booktitle = {2020 {IEEE/CVF} Conference on Computer Vision and Pattern Recognition,
                  {CVPR} 2020, Seattle, WA, USA, June 13-19, 2020},
	pages = {1465--1474},
	year = {2020}
}

@article{DBLP:journals/corr/abs-2407-01054,
	title = {Joint Pruning and Channel-wise Mixed-precision Quantization for Efficient Deep Neural Networks},
	author = {B. Alessandra Motetti and M. Risso and A. Burrello and E. Macii and M. Poncino and D. Jahier Pagliari},
	journal = {arXiv:2407.01054},
	year = {2024}
}

@article{DBLP:journals/tist/RokhAK23,
	title = {A Comprehensive Survey on Model Quantization for Deep Neural Networks in Image Classification},
	author = {B. Rokh and A. Azarpeyvand and A. Khanteymoori},
	journal = {{ACM} Trans. Intell. Syst. Technol.},
	volume = {14},
	number = {6},
	pages = {97:1--97:50},
	year = {2023}
}

@article{bouza2023estimate,
	title = {How to Estimate Carbon Footprint When Training Deep Learning Models? A Guide and Review},
	author = {L. Bouza and A. Bugeau and L. Lannelongue},
	journal = {Environmental Research Communications},
	volume = {5},
	number = {11},
	pages = {115014},
	year = {2023}
}

@inproceedings{DBLP:conf/icml/TanL21,
	title = {{EfficientNetV2}: Smaller Models and Faster Training},
	author = {M. Tan and Q. Le},
	editor = {M. Meila and T. Zhang},
	booktitle = {Proceedings of the 38th International Conference on Machine Learning,
                  {ICML} 2021, 18-24 July 2021, Virtual Event},
	volume = {139},
	pages = {10096--10106},
	year = {2021}
}

@inproceedings{DBLP:conf/eccv/TianKI20,
	title = {Contrastive Multiview Coding},
	author = {Y. Tian and D. Krishnan and P. Isola},
	editor = {A. Vedaldi and H. Bischof and T. Brox and J. Frahm},
	booktitle = {Computer Vision - {ECCV} 2020 - 16th European Conference, Glasgow,
                  UK, August 23-28, 2020, Proceedings, Part {XI}},
	volume = {12356},
	pages = {776--794},
	year = {2020}
}

@inproceedings{DBLP:conf/date/SchererCMB24,
	title = {Work in Progress: Linear Transformers for {TinyML}},
	author = {M. Scherer and C. Cioflan and M. Magno and L. Benini},
	booktitle = {Design, Automation {\&} Test in Europe Conference {\&} Exhibition,
                  {DATE} 2024, Valencia, Spain, March 25-27, 2024},
	pages = {1--2},
	year = {2024}
}

@article{mauricio2023comparing,
	title = {Comparing Vision Transformers and Convolutional Neural Networks for Image Classification: A Literature Review},
	author = {J. Maur{\'\i}cio and I. Domingues and J. Bernardino},
	journal = {Applied Sciences},
	volume = {13},
	number = {9},
	pages = {5521},
	year = {2023}
}

@article{you2020shiftaddnet,
	title = {Shiftaddnet: A Hardware-inspired Deep Network},
	author = {H. You and X. Chen and Y. Zhang and C. Li and S. Li and Z. Liu and Z. Wang and Y. Lin},
	journal = {Advances in Neural Information Processing Systems},
	volume = {33},
	pages = {2771--2783},
	year = {2020}
}
%\bibliography{tmlr}

\clearpage
\appendix
\section{Appendix} \label{appendix}

\subsection{Optimization Results}\label{appendix tables}
We provide tables that list the Pareto-optimal configurations obtained for ResNet20, MobileNet, GoogLeNet and EfficientNetV2 on CIFAR-10 and Caltech101 (Tables \ref{table resnet cifar} to \ref{table efficientnet cifar}). For each model and dataset, we report the mean and standard deviation across multiple seeds. These results include both the Pareto-optimal solutions and the performance of the default configurations in terms of loss and emissions, offering a comprehensive overview of the trade-offs achieved.

\subsection{Hyperparameter Importances}\label{appendix importances}
\begin{figure}[ht]
    \centering
    \begin{subfigure}[t]{\textwidth}
        \centering
        \includegraphics[width=\textwidth]{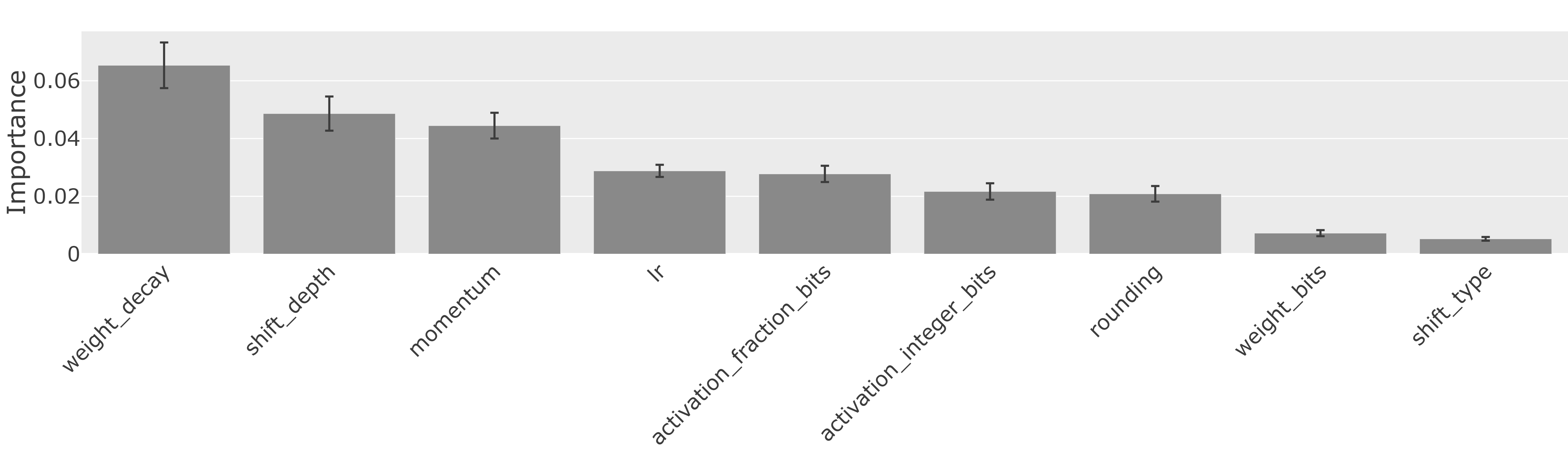}
        \caption{Hyperparameter importance with respect to loss.}
        \label{fig:importances_loss_mobilenet_caltech}
    \end{subfigure}
    
    % \vspace{0.5cm} % Optional vertical space between the subfigures
    
    \begin{subfigure}[t]{\textwidth}
        \centering
        \includegraphics[width=\textwidth]{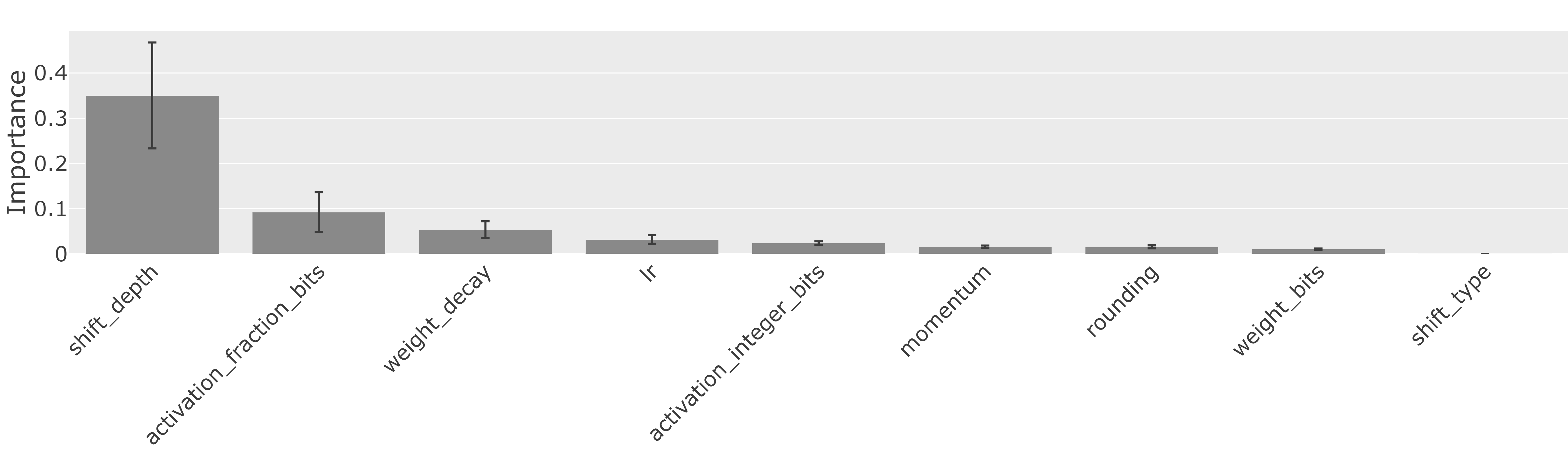}
        \caption{Hyperparameter importance with respect to emissions.}
        \label{fig:importances_emissions_mobilenet_caltech}
    \end{subfigure}
    
    \captionsetup{justification=centering}
    \caption{Hyperparameter importance according to fANOVA for MobileNet on Caltech101. (a) Importance with respect to loss. (b) Importance with respect to emissions.}
    \label{fig:fanova_importances mobilenet cifar}
\end{figure}

\begin{figure}[ht]
    \centering
    \begin{subfigure}[t]{\textwidth}
        \centering
        \includegraphics[width=\textwidth]{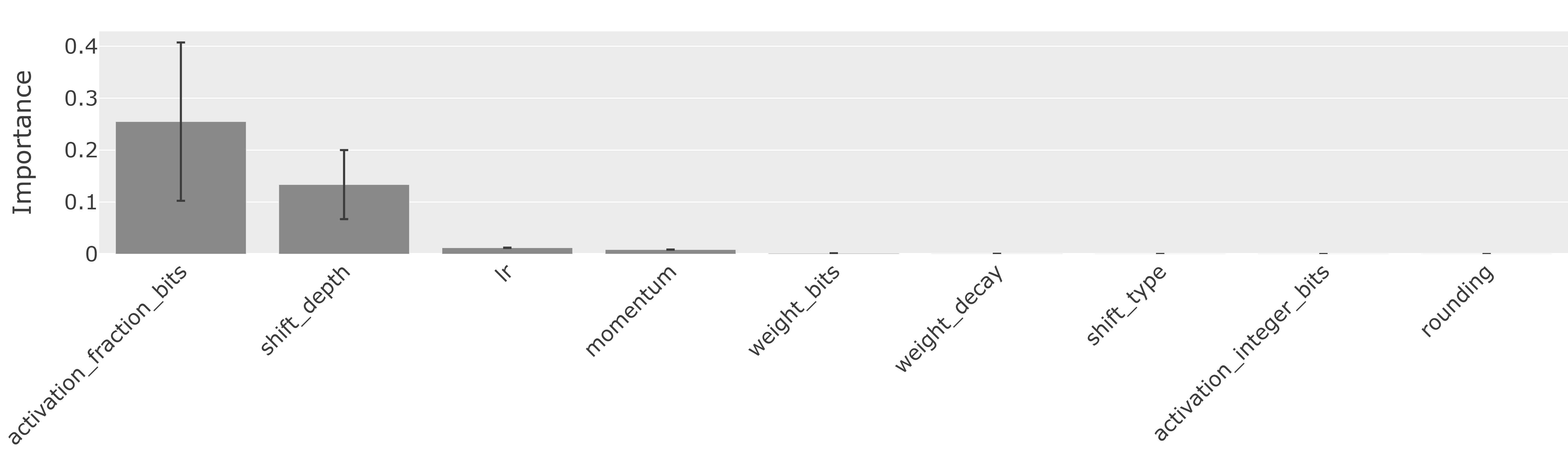}
        \caption{Hyperparameter importance with respect to loss.}
        \label{fig:importances_loss_googlenet_caltech}
    \end{subfigure}
    
    % \vspace{0.5cm} % Optional vertical space between the subfigures
    
    \begin{subfigure}[t]{\textwidth}
        \centering
        \includegraphics[width=\textwidth]{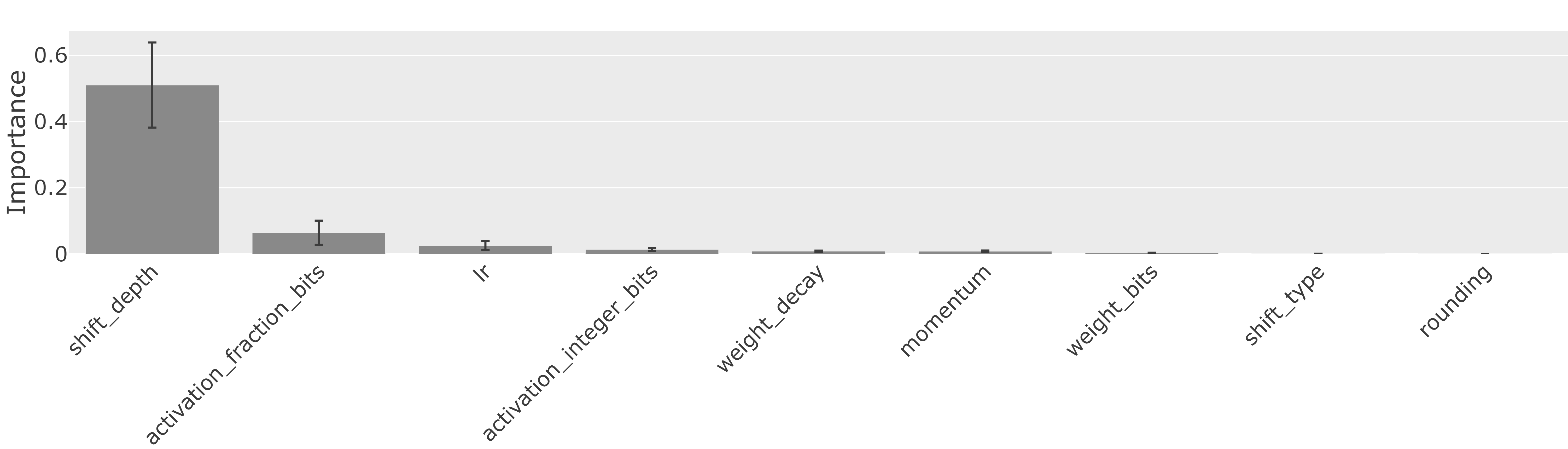}
        \caption{Hyperparameter importance with respect to emissions.}
        \label{fig:importances_emissions_googlenet_caltech}
    \end{subfigure}
    
    \captionsetup{justification=centering}
    \caption{Hyperparameter importance according to fANOVA for GoogLeNet on Caltech101. (a) Importance with respect to loss. (b) Importance with respect to emissions.}
    \label{fig:fanova_importances googlenet caltech}
\end{figure}

\begin{figure}[ht]
    \centering
    \begin{subfigure}[t]{\textwidth}
        \centering
        \includegraphics[width=\textwidth]{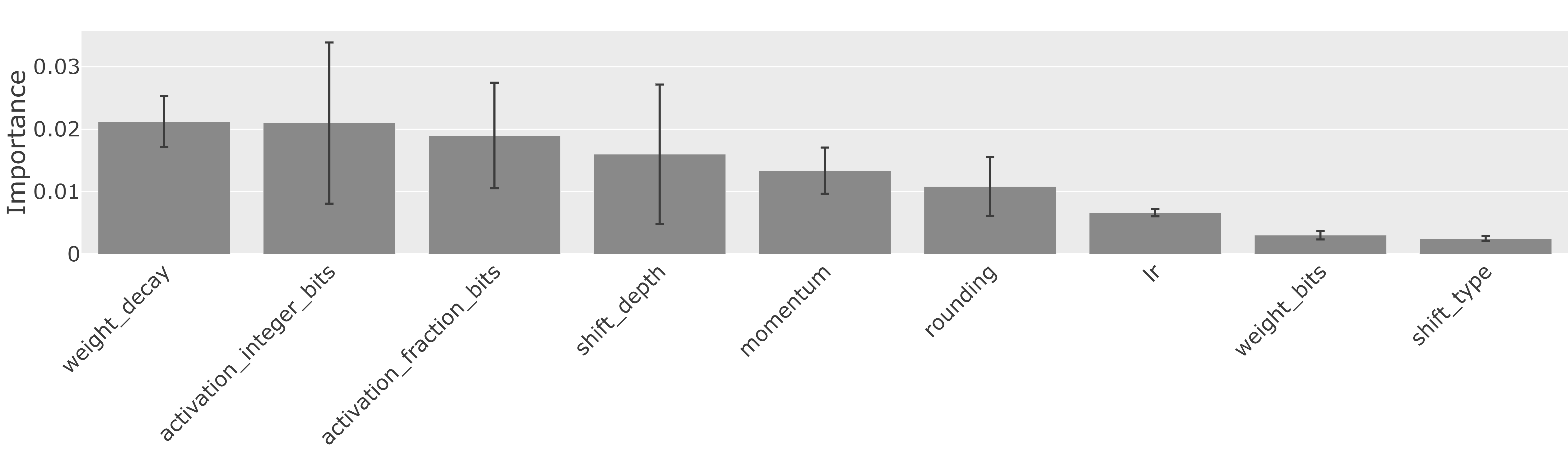}
        \caption{Hyperparameter importance with respect to loss.}
        \label{fig:importances_loss_resnet_caltech}
    \end{subfigure}
    
    % \vspace{0.5cm} % Optional vertical space between the subfigures
    
    \begin{subfigure}[t]{\textwidth}
        \centering
        \includegraphics[width=\textwidth]{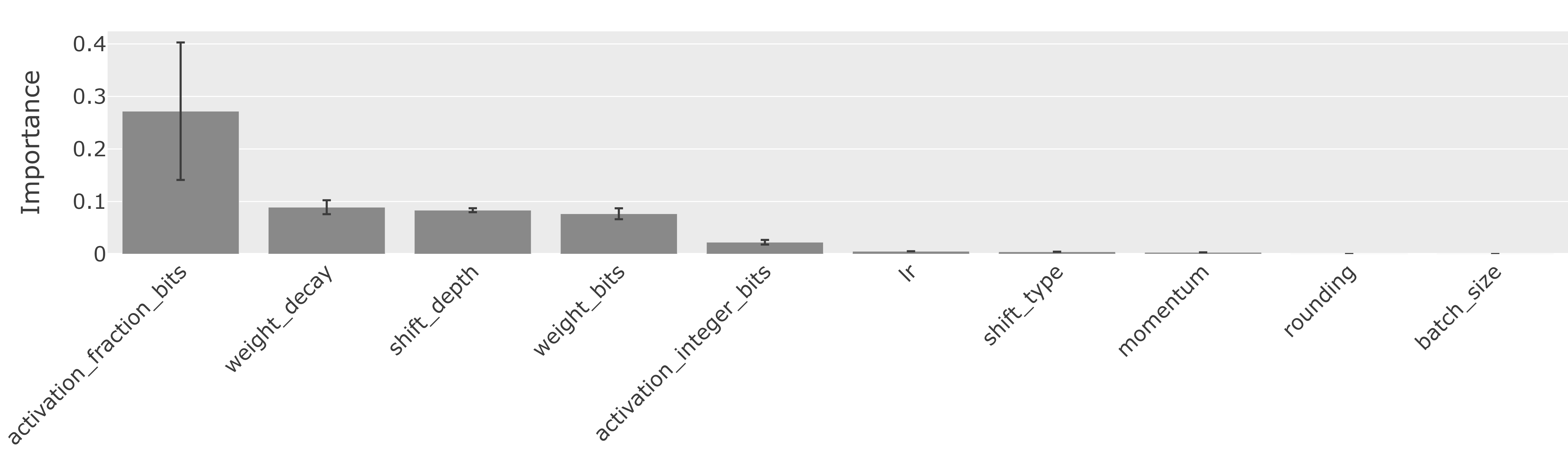}
        \caption{Hyperparameter importance with respect to emissions.}
        \label{fig:importances_emissions_resnet_caltech}
    \end{subfigure}
    
    \captionsetup{justification=centering}
    \caption{Hyperparameter importance according to fANOVA for ResNet20 on Caltech101. (a) Importance with respect to loss. (b) Importance with respect to emissions.}
    \label{fig:fanova_importances resnet caltech}
\end{figure}

\begin{figure}[ht]
    \centering
    \begin{subfigure}[t]{\textwidth}
        \centering
        \includegraphics[width=\textwidth]{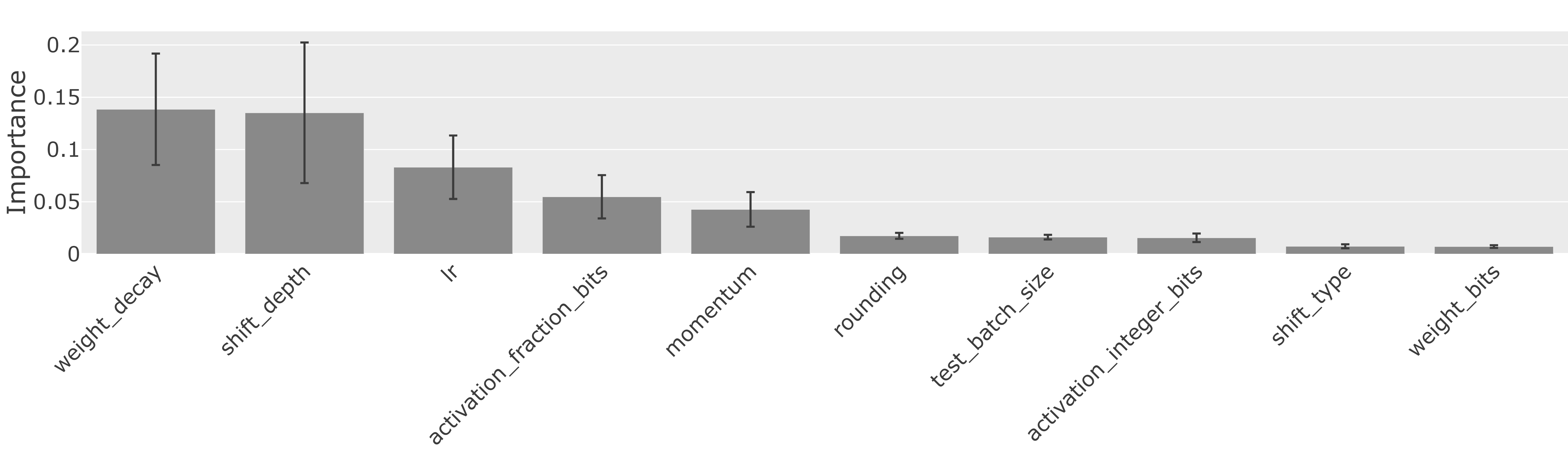}
        \caption{Hyperparameter importance with respect to loss.}
        \label{fig:importances_loss_mobilenet_cifar}
    \end{subfigure}
    
    \vspace{0.5cm} % Optional vertical space between the subfigures
    
    \begin{subfigure}[t]{\textwidth}
        \centering
        \includegraphics[width=\textwidth]{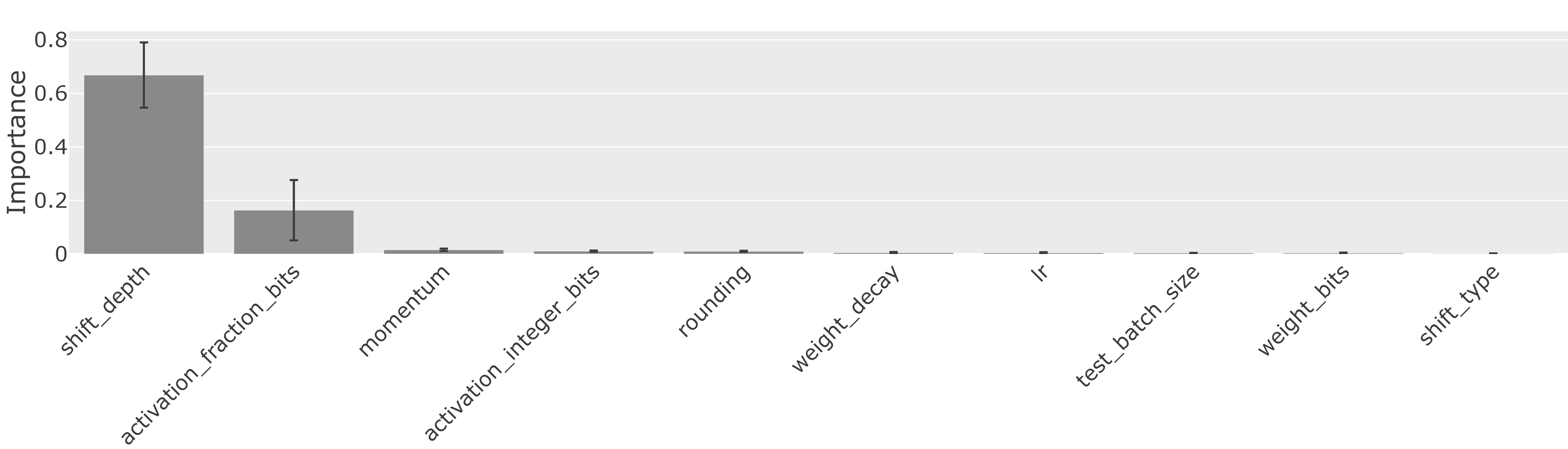}
        \caption{Hyperparameter importance with respect to emissions.}
        \label{fig:importances_emissions_mobilenet_cifar}
    \end{subfigure}
    
    \captionsetup{justification=centering}
    \caption{Hyperparameter importance according to fANOVA for MobileNet on CIFAR10. (a) Importance with respect to loss. (b) Importance with respect to emissions.}
    \label{fig:fanova_importances mobilenet cifar}
\end{figure}

\begin{figure}[ht]
    \centering
    \begin{subfigure}[t]{\textwidth}
        \centering
        \includegraphics[width=\textwidth]{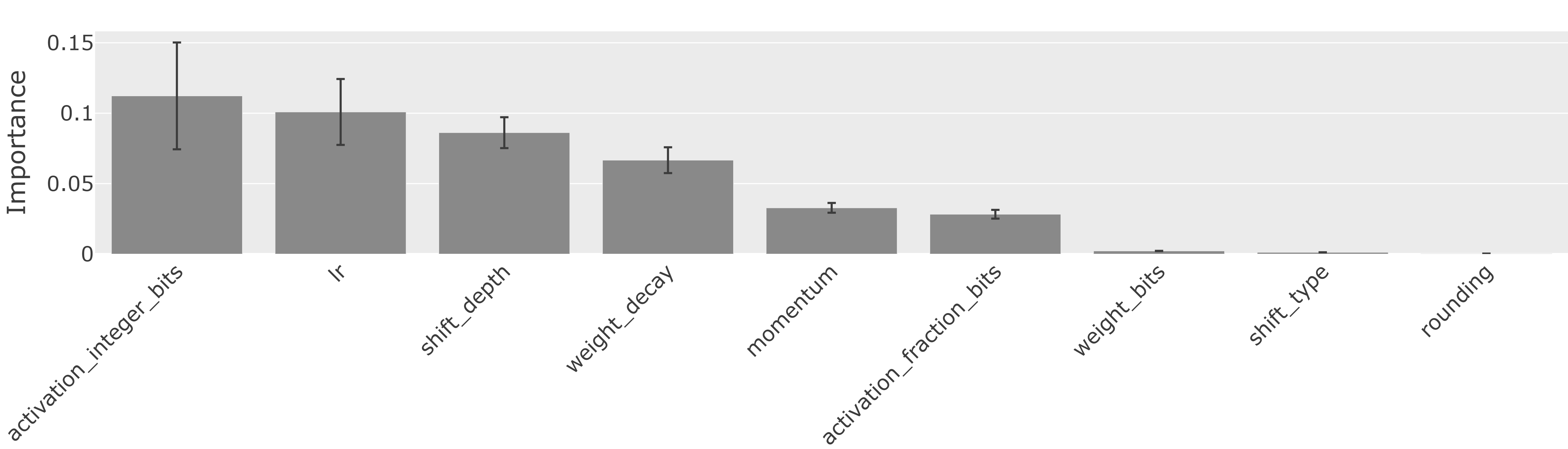}
        \caption{Hyperparameter importance with respect to loss.}
        \label{fig:importances_loss_googlenet_cifar}
    \end{subfigure}
    
    \vspace{0.5cm} % Optional vertical space between the subfigures
    
    \begin{subfigure}[t]{\textwidth}
        \centering
        \includegraphics[width=\textwidth]{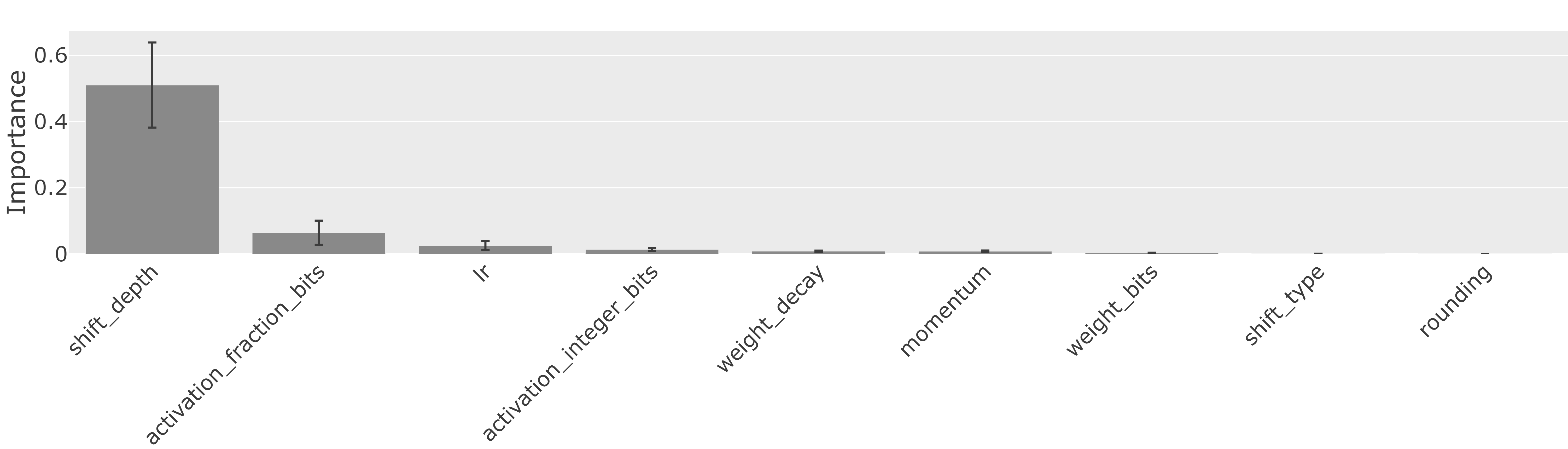}
        \caption{Hyperparameter importance with respect to emissions.}
        \label{fig:importances_emissions_googlenet_cifar}
    \end{subfigure}
    
    \captionsetup{justification=centering}
    \caption{Hyperparameter importance according to fANOVA for GoogLeNet on CIFAR10. (a) Importance with respect to loss. (b) Importance with respect to emissions.}
    \label{fig:fanova_importances googlenet cifar}
\end{figure}

We further analyzed the hyperparameter importance for the additional architectures we examined, as shown in Figures \ref{fig:importances_loss_mobilenet_caltech} to \ref{fig:importances_emissions_googlenet_cifar}. Across all models, the DSNN-specific hyperparameters consistently remain among the most important ones, alongside weight decay and learning rate. This highlights the relevance of our approach: focusing on the DSNN-specific parameters and allocating computational resources toward tuning them optimally for the given task appears to be a promising and effective strategy.

\subsection{Additional Baselines for Comparison}

To demonstrate the effectiveness and relevance of our approach, we compare against the same baselines as the original DSNN paper by \cite{DBLP:conf/cvpr/ElhoushiCSTL21}.
Since our work builds on a multi-objective optimization framework targeting both accuracy and emissions, but the original DSNN paper provides only single-objective baselines (focused solely on accuracy), we restricted our evaluation to the accuracy objective for fair comparison. Specifically, we focus on the VGG19 architecture \citep{simonyan-arxiv14a}, as this is the network for which \cite{DBLP:conf/cvpr/ElhoushiCSTL21} published baseline results.

From the Pareto front obtained via MOMF optimization of the VGG19 architecture, we selected a Pareto-optimal configuration that outperformed the default in accuracy while maintaining similar emissions. As shown in Table \ref{tab:vgg19small}, the final accuracy of this configuration outperforms not only the original unquantized VGG19 but also the quantized network baselines such as the DeepShift-PS baseline from \cite{DBLP:conf/cvpr/ElhoushiCSTL21}, AdderNet \citep{chen-arxiv20a}, and ShiftAddNet \citep{you2020shiftaddnet}.

\begin{table}[h]
\centering
\begin{tabular}{l l r}
\hline
& \textbf{Model} & \textbf{Accuracy} \\
\hline
\multirow{4}{*}{\textbf{VGG19 on CIFAR10}} 
& \textbf{AutoDSNN}      & \textbf{93.45\%} \\
& Original*          & 92\% \\
& DeepShift-PS* & 91.57\%\\
& AdderNet*~\citep{chen-arxiv20a}    & 93.02\% \\
& ShiftAddNet*~\citep{you2020shiftaddnet} & 90\% \\
\hline
\end{tabular}
\caption{Accuracy comparison of VGG19 model variants on CIFAR10. Baseline results from \cite{DBLP:conf/cvpr/ElhoushiCSTL21} are marked with an asterisk.}
\label{tab:vgg19small}
\end{table}

Looking at the configuration details in Table \ref{table:vgg19_config32}, we can see that the Pareto optimal configuration that beats the aforementioned baselines has 11 shift layers, meaning that little more than half of the VGG19 layers are quantized. This is in line with our analysis of optimal DSNN configurations in Section \ref{insights subsection}.
Again, it is not optimal to simply quantize the whole architecture.
More so, since with our MOMF approach we can beat both the original unquantized architecture, the default DSNN, as well as both quantization baselines.
These results highlight the relevance of our approach to discovering well-performing DeepShift configurations tailored to both accuracy and emissions. In particular, our optimal configuration, which outperforms all considered baselines, is only partially quantized, demonstrating that effective configurations are not straightforward or easily designed by hand. This underscores the importance of our method for systematically identifying efficient DSNNs beyond standard quantization heuristics, which are competitive with relevant baselines.

\begin{table}[ht]
\centering
\caption{Configuration for VGG19 on CIFAR10, selected from the Pareto front. This configuration achieves a significant improvement over the default unquantized VGG19 in terms of accuracy.}
\label{table:vgg19_config32}
\footnotesize
\setlength{\tabcolsep}{2pt}
\renewcommand{\arraystretch}{1.2}
\begin{tabularx}{0.7\columnwidth}{@{}X>{\centering\arraybackslash}X@{}}
\toprule
\textbf{Hyperparameter} & \textbf{Config 32} \\
\midrule
Optimizer & Adam \\
Learning Rate & 0.0107 \\
Momentum & 0.1474 \\
Weight Decay & 0.0032 \\
\midrule
Weight Bits & 5 \\
Act. Int. Bits & 20 \\
Act. Frac. Bits & 11 \\
Shift Depth & 11 \\
Shift Type & Q \\
Rounding & Stochastic \\
\midrule
\end{tabularx}
\end{table}

\begin{table}[ht]
\centering
\caption{Pareto optimal configurations and default DSNN initiation of ResNet 20 on CIFAR10. Pareto optimal solutions on the aggregated Pareto front of the ResNet20 architecture on CIFAR10 on three seeds, including the mean aggregated loss and emissions of the default configuration.}
\label{table resnet cifar}
\footnotesize % Reduces the font size
\setlength{\tabcolsep}{2pt} % Adjust the space between columns
\renewcommand{\arraystretch}{1.2} % Adjust vertical spacing
\begin{tabularx}{\columnwidth}{@{}X>{\centering\arraybackslash}X>{\centering\arraybackslash}X>{\centering\arraybackslash}X>{\centering\arraybackslash}X>{\centering\arraybackslash}X>{\centering\arraybackslash}X>{\centering\arraybackslash}X>{\centering\arraybackslash}X@{}} % Center alignment with automatic width adjustment
\toprule
\textbf{Hyperpar.} & \textbf{Config 124} & \textbf{Config 116} & \textbf{Config 71} & \textbf{Config 38} & \textbf{Config 82} & \textbf{Config 28} & \textbf{Config 49} & Default \\
\midrule
Optimizer & Ranger & Ranger & Ranger & Ranger & Ranger & Adagrad & Adagrad & SGD \\
Learning Rate & 0.0130 & 0.0327 & 0.0150 & 0.0542 & 0.0129 & 0.0548 & 0.0929 & 0.1 \\
Momentum & 0.7489 & 0.7718 & 0.1529 & 0.4983 & 0.6838 & 0.6783 & 0.4825 & 0.9 \\
Weight \\Decay & 0.0001 & 0.00005 & 0.0038 & 0.0001 & 0.0001 & 0.0002 & 0.0003 & 0.0001 \\
\midrule
Weight Bits & 2 & 8 & 5 & 2 & 2 & 5 & 5 & 5 \\
Act. Integer Bits & 11 & 13 & 2 & 9 & 11 & 22 & 24 & 16 \\
Act. Fraction Bits & 32 & 30 & 32 & 11 & 27 & 11 & 8 & 16 \\
Shift Depth & 1 & 1 & 3 & 1 & 1 & 1 & 1 & 20 \\
Shift Type & PS & PS & Q & PS & PS & PS & PS & PS \\
Rounding & Det.& Det. & Det. & Stochastic & Det. & Stochastic & Stochastic & Det. \\
\midrule
\textbf{Loss}&0.1127 & 0.1142 & 0.1161 & 0.1172 & 0.1176& 0.1352 & 0.1443&0.3518\\
\textbf{Emissions}& 0.000798 & 0.000792 & 0.000759 & 0.000708 & 0.000699 & 0.000682 & 0.000674& 0.000751\\
\bottomrule
\end{tabularx}
\end{table}

\begin{table}[ht]
\centering
\caption{Pareto optimal configurations and default DSNN instantiation of MobileNetV2 on CIFAR10}
\label{table mobilenet cifar}
\footnotesize % Reduces the font size
\setlength{\tabcolsep}{2pt} % Adjust the space between columns
\renewcommand{\arraystretch}{1.2} % Adjust vertical spacing
\begin{tabularx}{\columnwidth}{@{}X>{\centering\arraybackslash}X>{\centering\arraybackslash}X>{\centering\arraybackslash}X>{\centering\arraybackslash}X>{\centering\arraybackslash}X@{}}
\toprule
\textbf{Hyperparameter} & \textbf{Config 15} & \textbf{Config 68} & \textbf{Config 66} & \textbf{Config 21} & \textbf{Default} \\
\midrule
Optimizer & Adadelta & Adadelta & Adadelta & SGD & SGD \\
Learning Rate & 0.182188 & 0.186665 & 0.197817 & 0.183219 & 0.1 \\
Momentum & 0.726835 & 0.689596 & 0.1837 & 0.76337 & 0.9 \\
Weight Decay & 0.002727 & 0.003048 & 0.00306 & 0.002277 & 0.0001 \\
\midrule
Weight Bits & 5 & 5 & 5 & 4 & 5 \\
Activation Integer Bits & 21 & 19 & 21 & 23 & 16 \\
Activation Fraction Bits & 31 & 31 & 32 & 16 & 16 \\
Shift Depth & 14 & 7 & 1 & 1 & 53 \\
Shift Type & PS & PS & PS & PS & PS \\
Rounding & Deterministic & Deterministic & Deterministic & Deterministic & Deterministic \\
\midrule
\textbf{Loss} & 0.1683 & 0.1756 & 0.1797 & 0.9016 & 0.3017 \\
\textbf{Emissions} & 0.000755 & 0.000655 & 0.000552 & 0.000549 & 0.001656 \\
\bottomrule
\end{tabularx}
\end{table}

\begin{table}[ht]
\centering
\caption{Pareto optimal configurations and default DSNN instantiation of GoogLeNet on CIFAR10}
\label{table googlenet cifar}
\footnotesize % Reduces the font size
\setlength{\tabcolsep}{2pt} % Adjust the space between columns
\renewcommand{\arraystretch}{1.2} % Adjust vertical spacing
\begin{tabularx}{\columnwidth}{@{}X>{\centering\arraybackslash}X>{\centering\arraybackslash}X>{\centering\arraybackslash}X>{\centering\arraybackslash}X>{\centering\arraybackslash}X>{\centering\arraybackslash}X@{}}
\toprule
\textbf{Hyperpar.} & \textbf{Config 65} & \textbf{Config 33} & \textbf{Config 28} & \textbf{Config 14} & \textbf{Config 32} & \textbf{Default} \\
\midrule
Optimizer & Adadelta & Adadelta & Ranger & Ranger & Adadelta & SGD \\
Learning Rate & 0.028997 & 0.023838 & 0.020002 & 0.058610 & 0.115941 & 0.1 \\
Momentum & 0.209328 & 0.494258 & 0.250184 & 0.673880 & 0.372339 & 0.9 \\
Weight Decay & 0.008487 & 0.006737 & 0.006691 & 0.002313 & 0.009464 & 0.0001 \\
\midrule
Weight Bits & 2 & 3 & 2 & 4 & 2 & 5 \\
Activation Integer Bits & 21 & 24 & 29 & 26 & 29 & 16 \\
Activation Fraction Bits & 4 & 5 & 4 & 8 & 7 & 16 \\
Shift Depth & 1 & 1 & 1 & 1 & 1 & 22 \\
Shift Type & Q & Q & Q & PS & PS & PS \\
Rounding & Stochastic & Deterministic & Deterministic & Deterministic & Deterministic & Deterministic \\
\midrule
\textbf{Loss} & 0.1347 & 0.1409 & 0.1636 & 0.1748 & 0.1850 & 0.1810 \\
\textbf{Emissions} & 0.000916 & 0.000912 & 0.000876 & 0.000840 & 0.000835 & 0.001388 \\
\bottomrule
\end{tabularx}
\end{table}

\begin{table}[ht]
\centering
\caption{Pareto optimal configurations and default DSNN instantiation of ResNet20 on Caltech101}
\label{table resnet caltech}
\footnotesize % Reduces the font size
\setlength{\tabcolsep}{2pt} % Adjust the space between columns
\renewcommand{\arraystretch}{1.2} % Adjust vertical spacing
\begin{tabularx}{\columnwidth}{@{}X>{\centering\arraybackslash}X>{\centering\arraybackslash}X>{\centering\arraybackslash}X>{\centering\arraybackslash}X>{\centering\arraybackslash}X@{}}
\toprule
\textbf{Hyperparameter} & \textbf{Config 26} & \textbf{Config 66} & \textbf{Config 32} & \textbf{Config 76} & \textbf{Default} \\
\midrule
Optimizer & Ranger & Ranger & Ranger & RMSProp & SGD \\
Learning Rate & 0.023 & 0.039 & 0.076 & 0.015 & 0.1 \\
Momentum & 0.559 & 0.333 & 0.344 & 0.551 & 0.9 \\
Weight Decay & 0.0033 & 0.0027 & 0.0029 & 0.0069 & 0.0001 \\
\midrule
Weight Bits & 2 & 5 & 2 & 5 & 5 \\
Activation Integer Bits & 24 & 21 & 22 & 25 & 16 \\
Activation Fraction Bits & 17 & 27 & 13 & 21 & 16 \\
Shift Depth & 2 & 2 & 1 & 1 & 20 \\
Shift Type & PS & PS & PS & Q & PS \\
Rounding & Deterministic & Stochastic & Stochastic & Deterministic & Deterministic \\
\midrule
\textbf{Loss} & 0.456 & 0.532 & 0.636 & 0.874 & 0.679 \\
\textbf{Emissions} & 0.00046 & 0.00044 & 0.00044 & 0.00044 & 0.00109 \\
\bottomrule
\end{tabularx}
\end{table}

\begin{table}[ht]
\centering
\caption{Pareto optimal configurations and default DSNN instantiation of MobileNetV2 on Caltech101}
\label{table mobilenet caltech}
\footnotesize % Reduces the font size
\setlength{\tabcolsep}{2pt} % Adjust the space between columns
\renewcommand{\arraystretch}{1.2} % Adjust vertical spacing
\begin{tabularx}{\columnwidth}{@{}X>{\centering\arraybackslash}X>{\centering\arraybackslash}X>{\centering\arraybackslash}X>{\centering\arraybackslash}X>{\centering\arraybackslash}X@{}}
\toprule
\textbf{Hyperparameter} & \textbf{Config 9} & \textbf{Config 65} & \textbf{Config 33} & \textbf{Config 74} & \textbf{Default} \\
\midrule
Optimizer & Adadelta & Adadelta & SGD & Adam & SGD \\
Learning Rate & 0.192 & 0.199 & 0.004 & 0.006 & 0.1 \\
Momentum & 0.510 & 0.545 & 0.016 & 0.012 & 0.9 \\
Weight Decay & 0.009 & 0.004 & 0.004 & 0.004 & 0.0001 \\
\midrule
Weight Bits & 3 & 2 & 5 & 4 & 5 \\
Activation Integer Bits & 8 & 12 & 11 & 32 & 16 \\
Activation Fraction Bits & 26 & 30 & 7 & 6 & 16 \\
Shift Depth & 6 & 3 & 1 & 1 & 53 \\
Shift Type & PS & PS & Q & Q & PS \\
Rounding & Deterministic & Deterministic & Deterministic & Stochastic & Deterministic \\
\midrule
\textbf{Loss} & 0.274 & 0.276 & 0.459 & 0.870 & 0.337 \\
\textbf{Emissions} & 0.00053 & 0.00049 & 0.00046 & 0.00046 & 0.00066 \\
\bottomrule
\end{tabularx}
\end{table}

\begin{table}[ht]
\centering
\caption{Pareto optimal configurations and default DSNN instantiation of GoogLeNet on Caltech101}
\label{googlenet caltech table}
\footnotesize % Reduces the font size
\setlength{\tabcolsep}{2pt} % Adjust the space between columns
\renewcommand{\arraystretch}{1.2} % Adjust vertical spacing
\begin{tabularx}{\columnwidth}{@{}X>{\centering\arraybackslash}X>{\centering\arraybackslash}X>{\centering\arraybackslash}X>{\centering\arraybackslash}X>{\centering\arraybackslash}X>{\centering\arraybackslash}X>{\centering\arraybackslash}X>{\centering\arraybackslash}X@{}}
\toprule
\textbf{Hyperpar.} & \textbf{Config 66} & \textbf{Config 25} & \textbf{Config 44} & \textbf{Config 63} & \textbf{Config 33} & \textbf{Config 21} & \textbf{Config 20} & \textbf{Default} \\
\midrule
Optimizer & Adadelta & Adadelta & Adadelta & Adagrad & Radam & Adagrad & Adam & SGD \\
Learning Rate & 0.058 & 0.059 & 0.052 & 0.13 & 0.199 & 0.187 & 0.017 & 0.1 \\
Momentum & 0.185 & 0.642 & 0.194 & 0.889 & 0.647 & 0.367 & 0.248 & 0.9 \\
Weight \\ Decay & 0.0004 & 0.0019 & 0.0029 & 0.0006 & 0.0027 & 0.0048 & 0.0059 & 0.0001 \\
\midrule
Weight Bits & 3 & 4 & 3 & 2 & 4 & 3 & 2 & 5 \\
Activation Integer Bits & 25 & 26 & 9 & 10 & 7 & 30 & 7 & 16 \\
Activation Fraction Bits & 5 & 8 & 19 & 7 & 20 & 4 & 30 & 16 \\
Shift Depth & 8 & 1 & 9 & 1 & 2 & 2 & 2 & 22 \\
Shift Type & Q & PS & PS & PS & PS & Q & Q & PS \\
Rounding & Det. & Det. & Det. & Det. & Det. & Stochastic & Stochastic & Det. \\
\midrule
\textbf{Loss} & 0.356 & 0.362 & 0.376 & 0.563 & 0.778 & 0.793 & 0.922 & 0.466 \\
\textbf{Emissions} & 0.0026 & 0.0026 & 0.0023 & 0.0022 & 0.0021 & 0.0021 & 0.0020 & 0.0027 \\
\bottomrule
\end{tabularx}
\end{table}

\begin{table}[ht]
\centering
\caption{Pareto optimal configurations and default DSNN instantiation of EfficientNetV2 on CIFAR10}
\label{table efficientnet cifar}
\footnotesize
\setlength{\tabcolsep}{2pt}
\renewcommand{\arraystretch}{1.2}
\begin{tabularx}{\columnwidth}{@{}X>{\centering\arraybackslash}X>{\centering\arraybackslash}X>{\centering\arraybackslash}X>{\centering\arraybackslash}X>{\centering\arraybackslash}X>{\centering\arraybackslash}X@{}}
\toprule
\textbf{Hyperpar.} & \textbf{Config 61} & \textbf{Config 75} & \textbf{Config 31} & \textbf{Config 71} & \textbf{Config 39} & \textbf{Default} \\
\midrule
Optimizer & RMSprop & RMSprop & Adadelta & Adam & Ranger & SGD \\
Learning Rate & 0.01737 & 0.03561 & 0.03508 & 0.17490 & 0.12583 & 0.1 \\
Momentum & 0.8861 & 0.6085 & 0.4682 & 0.3632 & 0.7668 & 0.9 \\
Weight Decay & 0.0000381 & 0.007015 & 0.003031 & 0.009879 & 0.007242 & 0.0001 \\
\midrule
Weight Bits & 5 & 2 & 3 & 4 & 2 & 5 \\
Activation Integer Bits & 8 & 15 & 22 & 16 & 9 & 16 \\
Activation Fraction Bits & 11 & 21 & 22 & 18 & 10 & 16 \\
Shift Depth & 1 & 51 & 48 & 28 & 42 & 342 \\
Shift Type & Q & PS & Q & PS & PS & PS \\
Rounding & Stochastic & Deterministic & Stochastic & Deterministic & Stochastic & Deterministic \\
\midrule
\textbf{Loss} & 0.1521 & 0.1539 & 0.1555 & 0.1653 & 0.1657 & 0.8283 \\
\textbf{Emissions} & 0.000973 & 0.000934 & 0.000926 & 0.000880 & 0.000870 & 0.002074 \\
\bottomrule
\end{tabularx}
\end{table}


\begin{figure}[ht]
  \centering
  \includegraphics[width=1\textwidth]{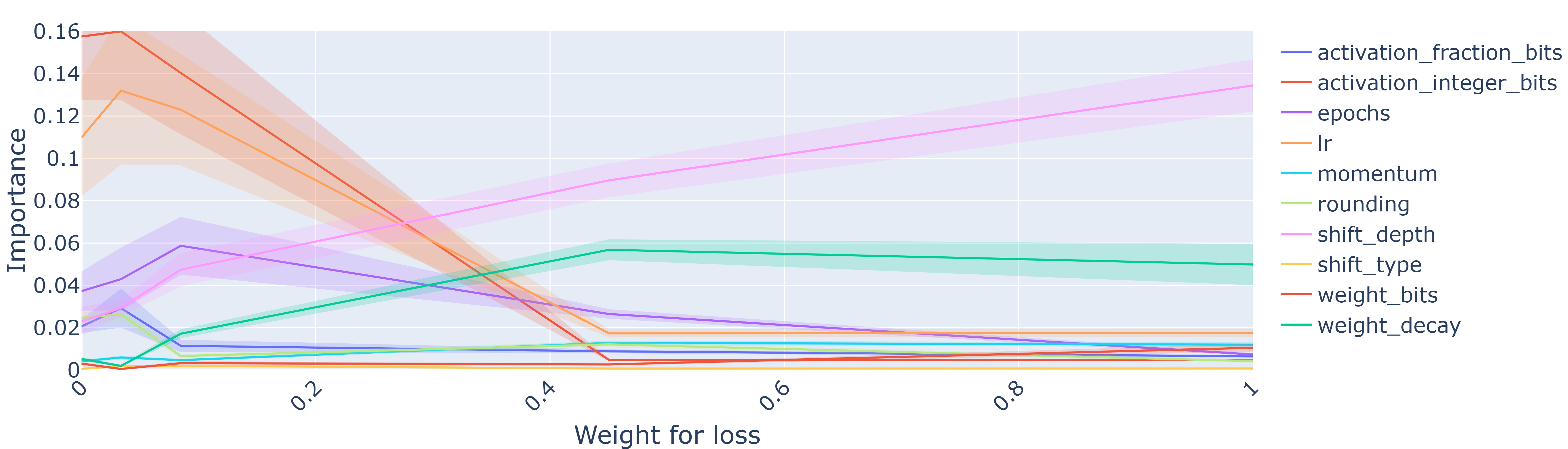}
  \caption{MO analysis of hyperparameter importance of a ResNet20 on CIFAR10 w.r.t. loss and emissions using the fANOVA method. The x-axis shows $w_l$, the weight of the objective loss, ranging from 0 to 1. The weight of the objective emissions is thus $w_e=1-w_l$. The y-axis shows the importance of each hyperparameter in the legend.}
  \label{fig:resnetcifar_mo_fanova}
\end{figure}

\clearpage
\begin{algorithm}[H]
\caption{Multi-Fidelity Optimization with Parego for DSNNs}
\label{algo:parego_dsnn}
\begin{algorithmic}[1]
\Require Configuration space $\mathcal{C}$, objectives $\mathcal{L} = [\mathcal{L}_\text{loss}, \mathcal{L}_\text{emissions}]$, 
         budget range $[B_{\min}, B_{\max}]$, number of trials $N$, 
         DSNN architecture $\mathcal{A}$.
\Ensure Pareto optimal configurations $\mathcal{P}_N$.

\State Initialize scenario $\mathcal{S}$ with $\mathcal{C}$, $\mathcal{L}$, $(B_{\min}, B_{\max})$, and $N$.
\State Generate initial observation dataset $\mathcal{D}_\text{init}$ by sampling $k$ random configurations $\{\lambda_i\}_{i=1}^k \subset \mathcal{C}$.
\State Define intensifier $\mathcal{H}$ as Hyperband for budget allocation.
\State Initialize optimizer $\mathcal{O}$ using $\text{Parego}$ and $\mathcal{H}$.

\For{each trial $t \in \{1, \ldots, N\}$}
    \State Select a configuration $\lambda_t \in \mathcal{C}$ using $\text{Parego}$.
    \State Allocate budget $b_t \in (B_{\min}, B_{\max})$ using $\mathcal{H}$.
    \State Perform evaluation of $\lambda_t$ with budget $b_t$:
    \begin{enumerate}
        \item Convert DSNN $\mathcal{A}$ to a shift-based architecture $\mathcal{A}'$:
        \begin{align*}
        \mathcal{A}' &= \text{convert\_to\_shift}(\mathcal{A}, \lambda_t[\text{shift\_depth}], \lambda_t[\text{shift\_type}]) \\
        w' &= \text{round}(w, \lambda_t[\text{rounding}]), \quad \forall w \in \mathcal{A}',
        \end{align*}
        where $\text{convert\_to\_shift}$ replaces standard operations with shift operations, 
        and $\text{round}$ applies deterministic or stochastic rounding.

        \item Train $\mathcal{A}'$ for $b_t$ epochs and compute the objective values:
        \begin{align*}
        \mathcal{L}_\text{loss}(\lambda_t) &= \frac{1}{|D_\text{test}|} \sum_{(x, y) \in D_\text{test}} \ell(f_{\mathcal{A}'}(x), y), \\
        \mathcal{L}_\text{emissions}(\lambda_t) &= \text{measure\_emissions}(\mathcal{A}', b_t),
        \end{align*}
        where $D_\text{test}$ is the test dataset, $\ell$ is the cross-entropy loss, and $\text{measure\_emissions}$ computes the energy consumption.

        \item Update observation dataset:
        \[
        \mathcal{D}_t \gets \mathcal{D}_{t-1} \cup \{(\lambda_t, [\mathcal{L}_\text{loss}(\lambda_t), \mathcal{L}_\text{emissions}(\lambda_t)])\}.
        \]
    \end{enumerate}
    \State Update the Pareto front:
    \[
    \mathcal{P}_t = \{(\lambda, \mathcal{L}(\lambda)) \in \mathcal{D}_t \mid \nexists \lambda' \in \mathcal{D}_t : \mathcal{L}(\lambda') \succ \mathcal{L}(\lambda)\},
    \]
    where $\mathcal{L}(\lambda') \succ \mathcal{L}(\lambda)$ denotes that $\lambda'$ dominates $\lambda$.

\EndFor
\State \Return $\mathcal{P}_N$
\end{algorithmic}
\end{algorithm}

\end{document}